\documentclass[10pt,twocolumn,letterpaper]{article}

\usepackage{cvpr}              

\usepackage[accsupp]{axessibility} 

\usepackage{algorithm}
\usepackage{algpseudocode} 
\usepackage{amsmath}
\usepackage{amsfonts}
\usepackage{amssymb}
\usepackage{siunitx}
\usepackage{multirow}
\usepackage{overpic}
\usepackage{subcaption}
\usepackage{tikz}
\usepackage{float}
\usetikzlibrary {angles,quotes}

\DeclareMathOperator*{\argmin}{argmin}

\definecolor{cvprblue}{rgb}{0.21,0.49,0.74}
\usepackage[pagebackref,breaklinks,colorlinks,citecolor=cvprblue]{hyperref}

\def\paperID{5538} 
\def\confName{CVPR}
\def\confYear{2024}

\title{Bayesian Differentiable Physics for Cloth Digitalization}

\author{Deshan Gong\\
University of Leeds\\
Leeds, United Kingdom\\
{\tt\small scdg@leeds.ac.uk}
\and
Ningtao Mao\\
University of Leeds\\
Leeds, United Kingdom\\
{\tt\small N.Mao@leeds.ac.uk}
\and
He Wang \footnotemark[1]\\
University College London\\
London, United Kingdom\\
{\tt\small he\_wang@ucl.ac.uk}
}

\begin{document}

\twocolumn[{
\maketitle
\begin{center}
    \captionsetup{type=figure}\addtocounter{figure}{-1}

    \begin{subfigure}[b]{\textwidth}
        \includegraphics[width=\textwidth]{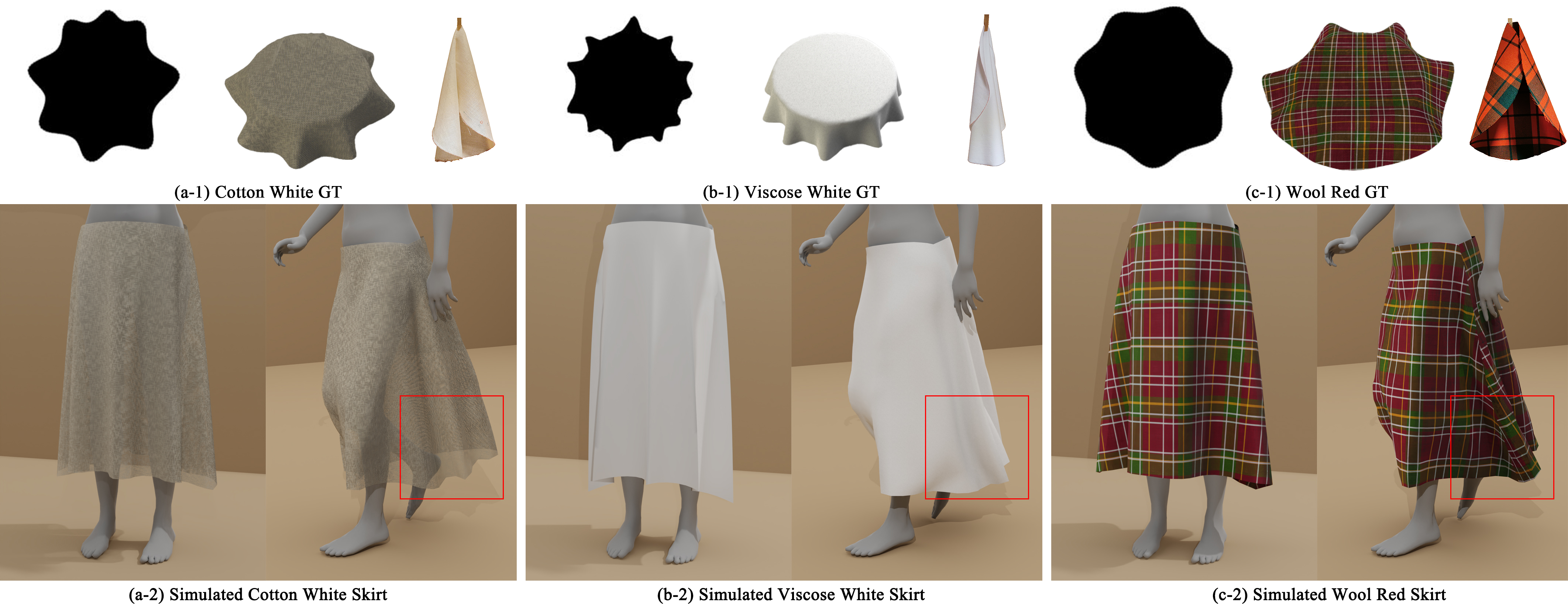}
    \end{subfigure}

    \captionof{figure}{We introduce a Bayesian Differentiable Physics (BPD) model for digitalizing real cloths by inferring their physical properties from the standard Cusick drape data (a-1, b-1, c-1 left). The digitalized cloths exhibit various drapabilities, faithfully reflecting their diverse mechanical characteristics and materials (a-1, b-1, c-1 middle and right). Further, our model enables the generalization of the learned mechanical characteristics and materials to garments (a-2, b-2, c-2).}
    \label{fig:teaser}
\end{center}
}]

\renewcommand{\thefootnote}{\fnsymbol{footnote}}
\footnotetext[1]{corresponding author, he\_wang@ucl.ac.uk}

\renewcommand{\thefootnote}{\arabic{footnote}}

\begin{abstract}
We propose a new method for cloth digitalization. Deviating from existing methods which learn from data captured under relatively casual settings, we propose to learn from data captured in strictly tested measuring protocols, and find plausible physical parameters of the cloths. However, such data is currently absent, so we first propose a new dataset with accurate cloth measurements. Further, the data size is considerably smaller than the ones in current deep learning, due to the nature of the data capture process. To learn from small data, we propose a new Bayesian differentiable cloth model to estimate the complex material heterogeneity of real cloths. It can provide highly accurate digitalization from very limited data samples. Through exhaustive evaluation and comparison, we show our method is \textbf{accurate} in cloth digitalization, \textbf{efficient} in learning from limited data samples, and \textbf{general} in capturing material variations. Code and data are available\footnote{\url{https://github.com/realcrane/Bayesian-Differentiable-Physics-for-Cloth-Digitalization}}
\end{abstract}    
\section{Introduction}
\label{sec:intro}

In the emergence of the Metaverse, being able to build digital replicas of specific real-world objects becomes highly desirable. Despite many efforts to digitalize relatively simple objects such as rigid bodies~\cite{strecke2021diffsdfsim}, human bodies~\cite{gartner2022differentiable}, challenges still remain for objects with complex behaviors such as cloth, fluid, and gases~\cite{li2022diffcloth, schenck2018spnets, xian2023fluidlab}. One particular challenge is to digitalize cloths, which can greatly benefit multiple application domains, \eg customized fashion design, computer animation, textile manufacturing, \etc.

The key to cloth digitalization is a model that can capture the complex physical behaviors of a given cloth (not cloths in general). Broadly speaking, existing research provides several potential avenues. Physics-based approaches employ physics models that explicitly represent the materials of cloths. Replicating a specific cloth then comes down to hand-tuning the material parameters~\cite{choi2005research} or solving an inverse problem~\cite{wang2011data,liang_2019_differentiable}. In parallel, in computer vision, data-driven approaches employ deep learning models to learn the physical behaviors from data, without explicit physics knowledge~\cite{kim2013near}. While the former requires laborious hand-tuning and slow optimization, which is still difficult to replicate a given cloth~\cite{bhat2003estimating}, the latter requires large amounts of data and still suffers from low accuracy when it comes to mimicking specific cloth samples~\cite{gundogdu2020garnet++,ncloth22}. Recently, a combination of physics and deep learning, \ie differentiable physics, provides a new direction~\cite{li2021diffcloth,liang_2019_differentiable,Gong_finegrained_2022}, but the ability to replicate the exact physical behaviors of a given cloth is still under-explored.

\begin{figure*}[tb]
    \centering
    \subcaptionbox{Tester \label{fig:testing}}{\includegraphics[width=0.16\textwidth]{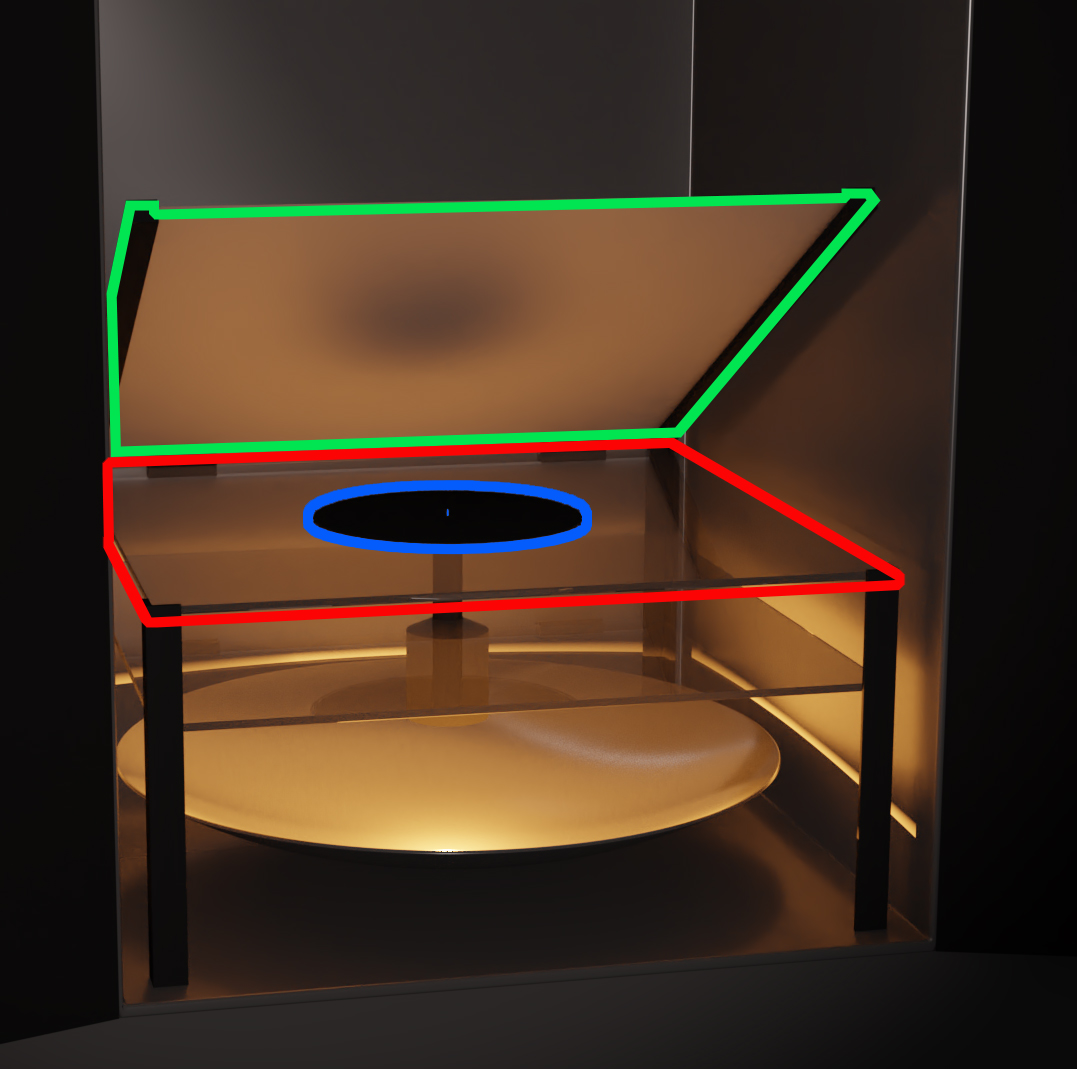}} \hfill
    \subcaptionbox{Initial State \label{fig:initial_state}}{\includegraphics[width=0.16\textwidth]{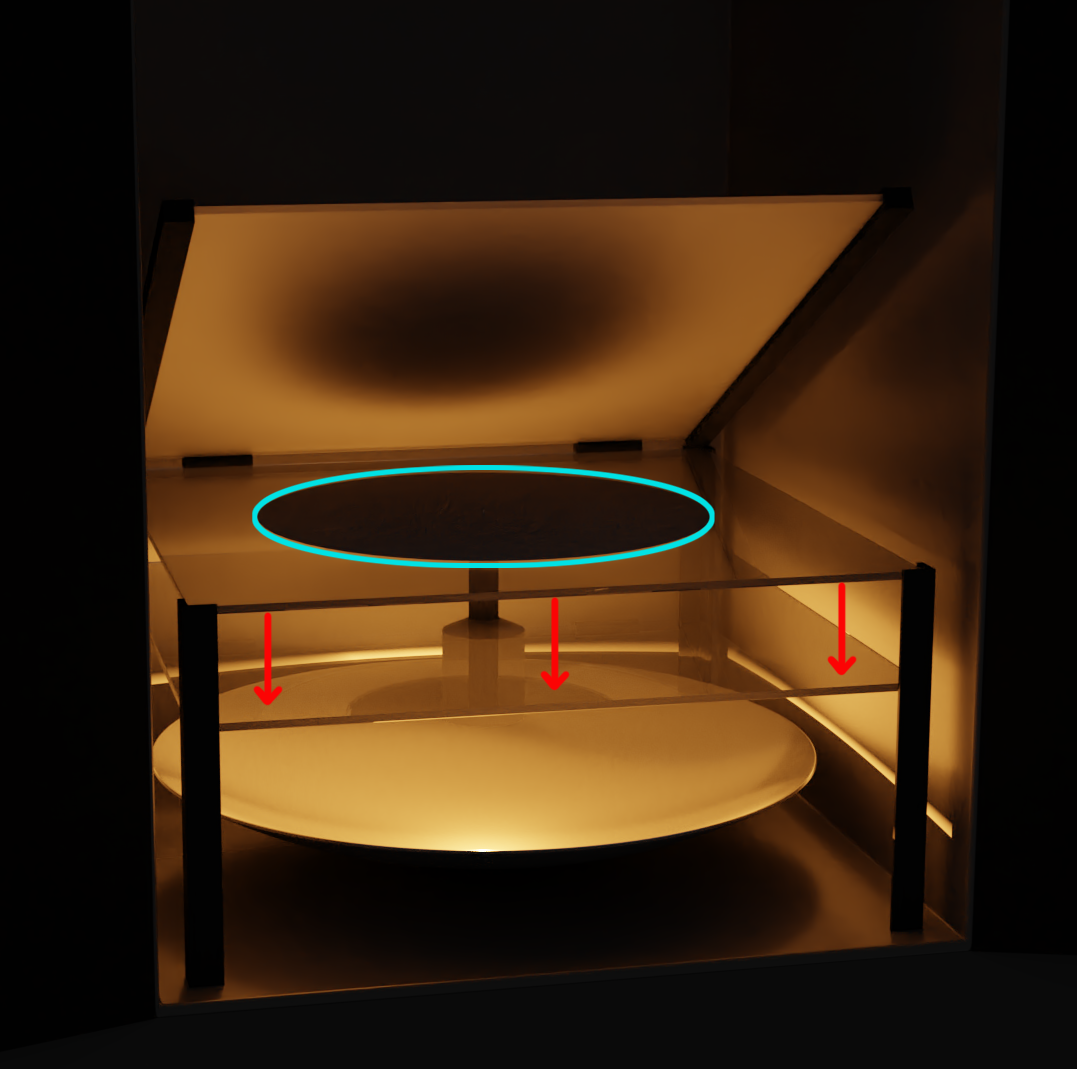}} \hfill
    \subcaptionbox{Drape State \label{fig:drape_state}}{\includegraphics[width=0.16\textwidth]{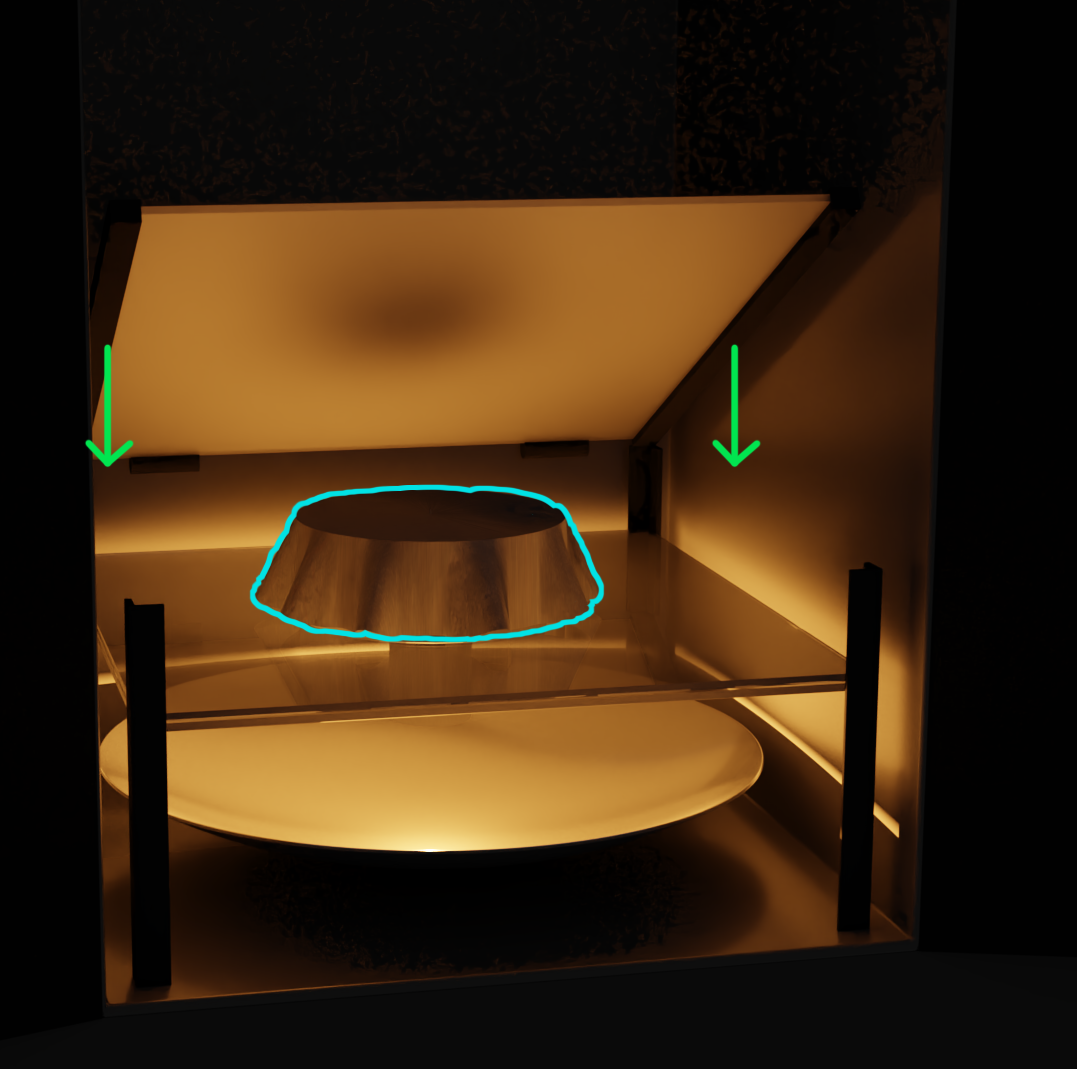}} \hfill
    \subcaptionbox{Capturing \label{fig:capturing}}{\includegraphics[width=0.16\textwidth]{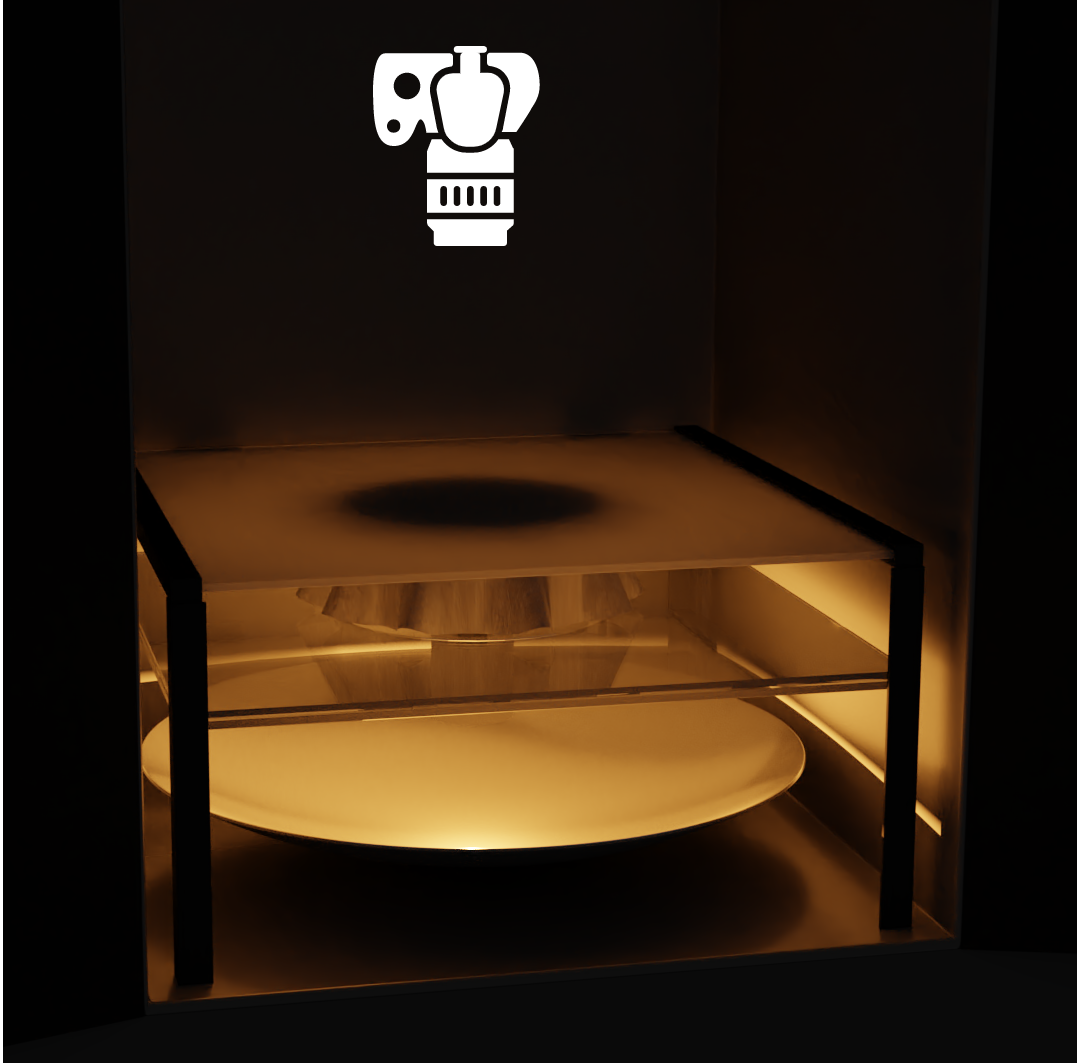}} \hfill
    \subcaptionbox{Raw Photo \label{fig:raw_photo}}{\includegraphics[width=0.16\textwidth]{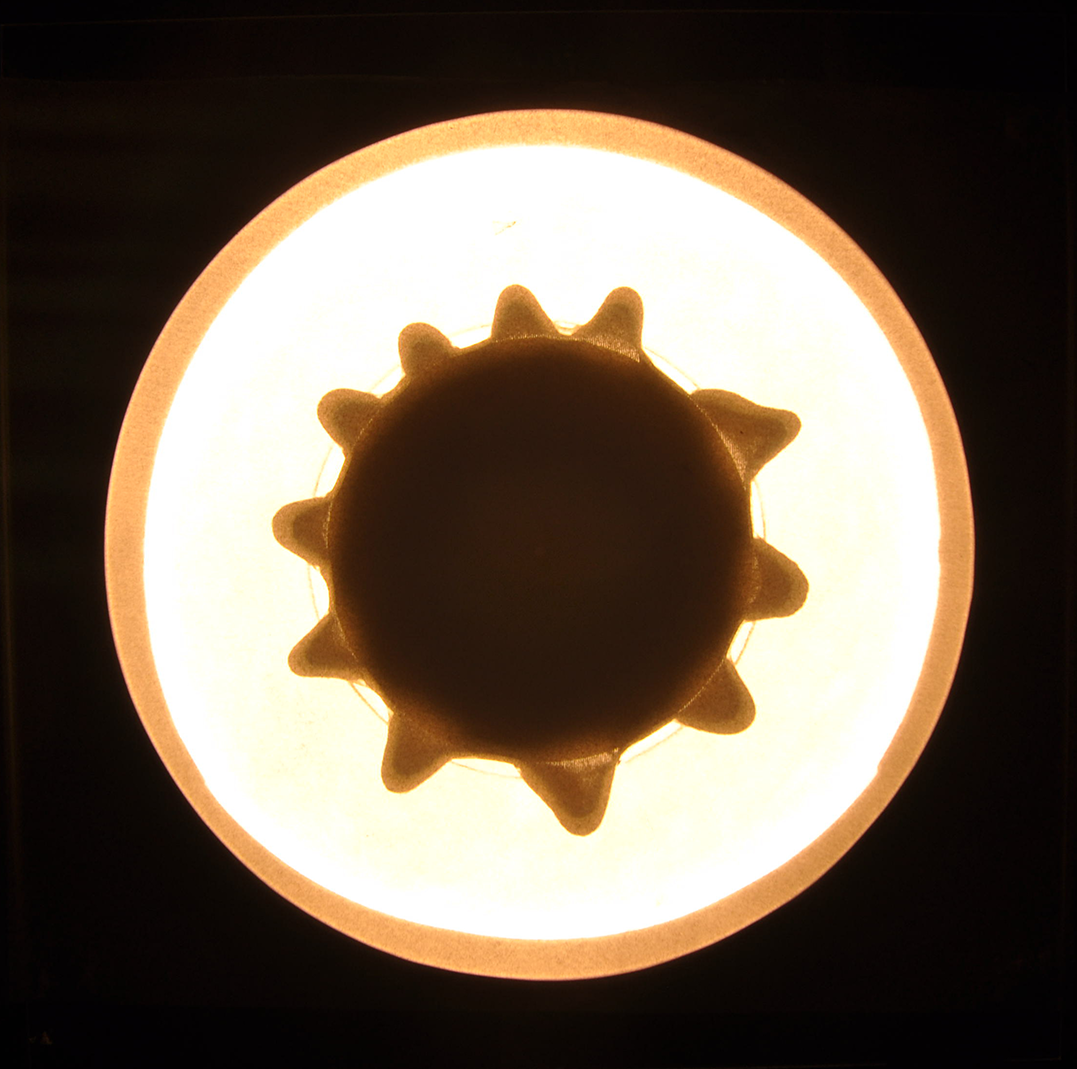}} \hfill
    \subcaptionbox{Silhouette \label{fig:silhouette}}{\includegraphics[width=0.16\textwidth]{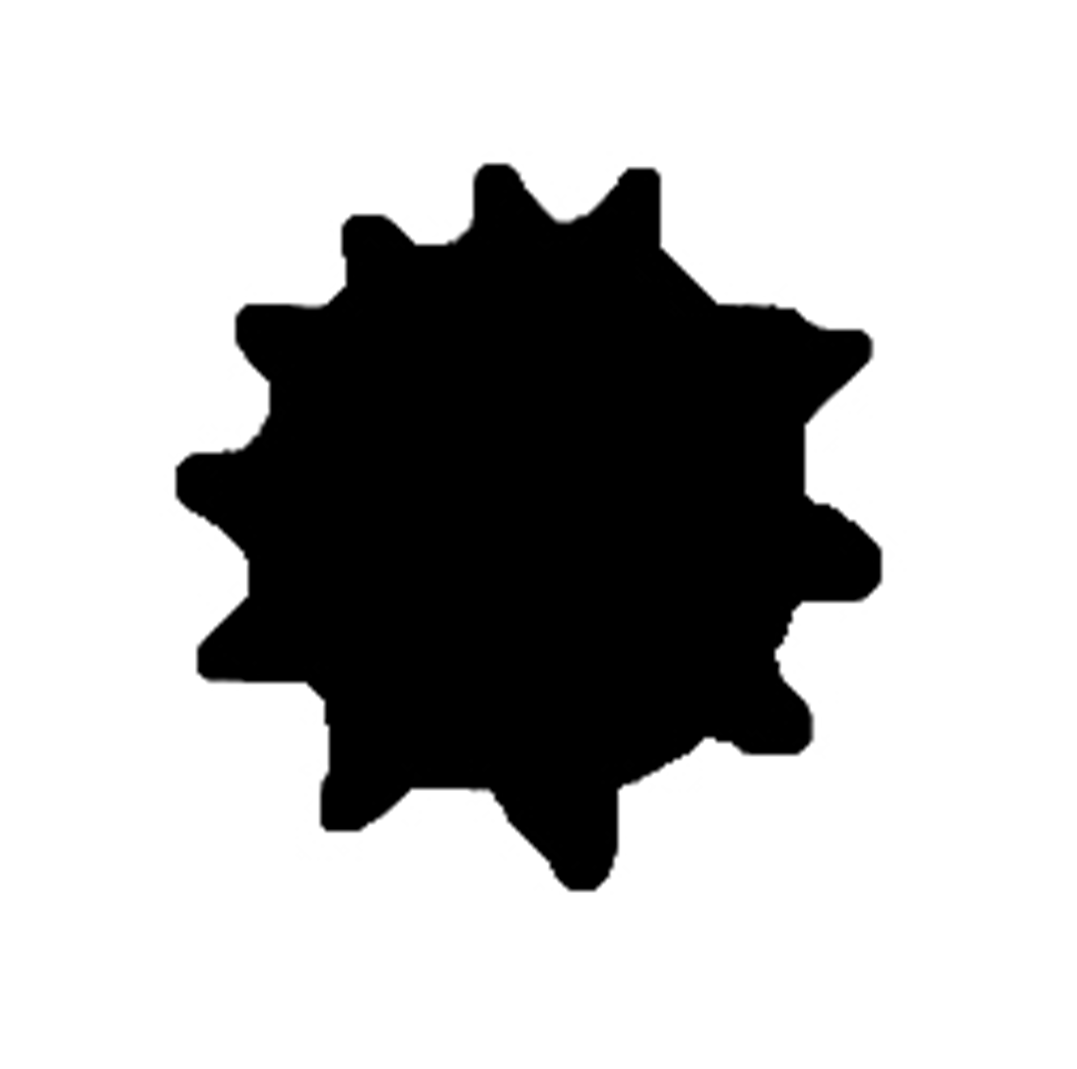}}
    \caption{Cusick drape testing. The Tester, \ie Cusick drape meter, has an inner support panel (blue), an outer support panel (red) and a frosted glass lid (green). A round cloth sample is first laid flat on the support panels (light blue in Initial State). Then the outer support panel is lowered to allow the cloth to naturally drape (Drape State). Next, the glass lid is closed so that the light source at the bottom can project the cloth to the lid which is recorded by a camera at the top (Capturing). Finally, the cloth Silhouette is extracted from the Raw Photo. The whole drape meter is in a black chamber so the testing process is not observable. Due to patent restrictions, the images are rendered, not the real device.}
\label{fig:Cusick}
\end{figure*}

We argue that the foremost challenge in digitalizing specific cloth samples is the measurement accuracy in data. Cloth materials and dynamics are subtle, calling for fine-controlled data capture, where there is a notable difference between current deep learning and the textile standards. Compared with textile where variables are strictly controlled, \eg temperature, air moisture, current deep learning learns from the data captured in far less controlled settings, \eg videos~\cite{bouman2013estimating, wang2011data}. Consequently the learned models are merely sufficient for general motion prediction/simulation \cite{yang2017learning, narain2012adaptive}, and are far from being accurate for detailed simulation, manufacturing, design~\cite{gao2021review, delavari2021mathematical, luible2008simulation}. We fill this gap by employing Cusick drape testing under the British Standard~\cite{cdi_bsi_primary_000000000030128826} (\cref{fig:Cusick}). At the high level, Cusick drape testing captures the cloth `drapability' in images and uses them to characterize the material. It is a vision-based approach which is machine learning friendly and effective in describing the cloth physical properties \cite{chu1950mechanics, cusick196130, cusick196546, cusick196821, collier1991measurement, collier1989effects, jeong1998studya, jeong1998studyb}. 

However, simply applying or adapting the existing methods~\cite{sanchez-gonzalez_learning_2020,liang_2019_differentiable,Gong_finegrained_2022,Li_diffcloth_2022} on Cusick drape data is difficult. First, while the latest differentiable physics methods often assume cloths as homogeneous materials~\cite{liang_2019_differentiable,li2021diffcloth, Gong_finegrained_2022, stuyck2023diffxpbd}, cloths are heterogeneous materials: the mechanical properties in different parts of the same sample are different. In addition, there exists dynamics stochasticity in cloth draping~\cite{pratihar2013static}. A deterministic and homogeneous model leads to averaged behaviors hence inaccurate digitalizations. Next, black-box deep learning methods cannot be used either, due to that they need a large amount of data~\cite{sanchez-gonzalez_learning_2020}. Collecting such data is prohibitively time-consuming and labor-intensive for accurate tests such as Cusick. This is why there are few public datasets of Cusick drape, compared with hours of videos at the disposal for deep learning. So we collect our own data, but the data size is not even remotely close to videos~\cite{runia2020cloth}. Finally, a challenge in learning from Cusick drape is that a standard tester (\ie Cusick drape meter) only provides one drape image (\cref{fig:Cusick}), \ie no 3D geometry or motion, ruling out the methods~\cite{sanchez-gonzalez_learning_2020,runia2020cloth} that require dynamics data.

To address the aforementioned data scarcity, dynamics stochasticity and material heterogeneity, we propose a new Bayesian learning scheme. Starting from the joint probability of a Cusick drape image and the initial state, we model the stochastic draping motion as a series of probabilistic state transitions, with the learnable physical parameters as latent variables. Due to data scarcity (i.e. one image per drape), inferring the latent variables is a formidable task. Therefore, we propose a new differentiable heterogeneous cloth model to govern the dynamics of the draping, and incorporate randomnesses in the material parameters to account for the draping stochasticity. Owning to its high sample efficiency, our model can learn from extremely limited data (\ie merely one image) of a draping sample. Furthermore, to account for the within-type material variations, we impose learnable 
posterior over the material parameters, leading to a new Bayesian differentiable cloth model, which can learn distributions of plausible physical parameters. Not only does it explicitly model the draping stochasticity, it also enables us to transfer the learned material to arbitrary geometries such as garments.

We show that our method is \textit{accurate} in replicating highly plausible cloth mechanical behaviors, \textit{efficient} in training with limited data, and \textit{general} in capturing material variations. Further, the digitalized cloths can be used to simulate garments made from different materials displaying distinguishable mechanical characteristics. Since there is no existing deep learning method designed for similar tasks, we compare our method with possible alternative solutions including different cloth models and optimization methods. Formally, our contributions include: (1) a new method for cloth digitalization based on limited Cusick drape data, (2) a new Bayesian differentiable cloth model to enable accurate digitalization, and (3) a new dataset collected from the Cusick drape testing.

\section{Related Work}

\label{sec:related_work}


\paragraph{Cloth Digitialization} aims to create digital replicas of specific cloth samples. Broadly speaking, there are three approaches: supervised learning, self-supervised learning, and physical parameter estimation. Supervised learning mimicks cloth dynamics by learning from cloth~\cite{gundogdu2020garnet++, patel2020tailornet}, which requires a huge amount of training data and often suffers from low generalizability in unseen environments or materials~\cite{wang2021gpu}. This is because this type of methods encodes little physics and entirely relies on data. By contrast, self-supervised learning explicitly considers cloth physics and does not rely on data~\cite{bertiche2020pbns, santesteban2022snug}. However, this method usually employs simplified physics models and therefore suffers from issues such as over-smoothing and penetrations that are difficult to solve~\cite{bertiche2022neural}. Also, both approaches are designed to capture general behaviors of cloths, rather than replicating specific cloth samples.

Physical parameter estimation aims to infer the cloth simulation parameters so that it can reproduce the dynamics of real cloths. Compared with the aforementioned approaches, this method is the closest to building the exact digital replicas of specific cloth samples. Both model-free and model-based methods are proposed in this approach. Model-free methods learn a mapping function between the observed cloths and the physical parameters without explicitly modeling any physics~\cite{yang2017learning, bouman2013estimating, ju2020estimating}. Model-based methods employ physics models and optimize the simulation parameters to minimize the difference between the simulation and the observations. While model-free methods are simple to implement, they are data demanding and collecting real cloth data is usually prohibitively time-consuming. By contrast, model-based methods have outstanding data efficiency~\cite{wang2011data,clyde2017modeling,miguel2013modeling}. Recently, one line of model-based methods called differentiable cloth simulation has demonstrated high learning accuracy and convergence speed~\cite{liang_2019_differentiable,li2021diffcloth, Gong_finegrained_2022}. These methods leverage fully differentiable cloth simulators and gradient-based optimization to estimate the parameters. However, they need to learn from 3D geometries which are usually difficult to accurately capture. \cite{jatavallabhula2021gradsim} uses a differentiable renderer~\cite{kato2020differentiable,liu2019soft} so that it can learn from 2D images. However, the digitalized cloths are still significantly different from the real ones. 

Our method falls into the category of differentiable cloth models. However, we argue that one common issue in existing research is the data accuracy. Cloth mechanical behaviors are affected by the environment, e.g. temperature and air moisture~\cite{bishop1996fabrics}. Therefore, the casual data collection setups employed in existing research do not actively control these factors. Further, since fabric testing has widely recognized standards~\cite{clyde2017modeling}, we argue standard apparatuses and protocols should be used. In addition to the data accuracy, we argue that the current differentiable cloth models are overly simplified. To be able to digitalize specific cloth samples, the physics model should explicitly consider material heterogeneity and behavioral stochasticity, due to the wide range of materials used in fabrics~\cite{saito2014macroscopic,gazzo2019characterisation}.

To resolve these problems, we introduce a new accurate drape dataset which is collected following widely acknowledged standards~\cite{cusick196130,cusick196546,cusick196821}. Moreover, we propose a novel Bayesian differentiable cloth simulator that can more accurately digitalize real cloth behaviors by modeling the material heterogeneity and dynamics stochasticity through Bayesian inference with outstanding data efficiency.

\paragraph{Physics-based Deep Learning.}  Our research can be seen as a part of recent attempts to leverage deep learning to solve differential equations, which has spiked research interests to address issues such as noise modeling, finite element mesh generation and high dimensionality~\cite{Karniadakis_datadriven_2021,lu2021deepxde,beck2020overview,meng2022physics}. Deep Neural Networks (DNNs) can learn to generate Finite Element meshes for steady state problems~\cite{Zhang_MeshingNet_2020,zhang_meshingnet3d_2021}. Also, they can be part of Partial Differential Equations (PDEs) for purposes such as reduced-order modeling~\cite{Shen_high_2021,han2018solving}, noise estimation~\cite{yang2021b} and differentiable simulation~\cite{Gong_finegrained_2022,holl2020phiflow,yue2022human,yue2023human}. Further, DNNs can replace PDEs completely in physics-informed neural networks (PINNs)~\cite{raissi2019physics,song2024loss} where the process of solving PDEs is replaced by inference on trained DNNs. Different from existing work, we propose a Bayesian differentiable physics model for fabrics to explicitly digitalize their stochastic mechanical properties.
\section{Methodology}

\subsection{Cusick Drape Test}

Our Cusick drape meter comes with a chamber within which there are two support panels and one frosted glass lid (\cref{fig:testing}). During testing, we first cut a cloth sample into a round shape (\cref{fig:Mesh}(a)) and pin its center to the center of the blue panel in \cref{fig:testing} (diameter is 18cm), shown in \cref{fig:initial_state}. Then we lower the transparent panel (the red panel in \cref{fig:testing}) and let the cloth naturally drape until the transparent panel does not contact with the cloth (\cref{fig:drape_state}). Finally, an image $\mathcal{I} \in \{pix\in\mathbb{Z}: 0 \leq pix \leq 255\}^{L \times L}$ (\cref{fig:raw_photo}) is taken by a DSLR camera from the top (\cref{fig:capturing}). To minimize possible external perturbations such light interference, the chamber is closed when capturing.

\subsection{A Bayesian Model for Cusick Drape}
\begin{figure}[tb]
    \centering
    \begin{overpic}[width=0.3\textwidth]{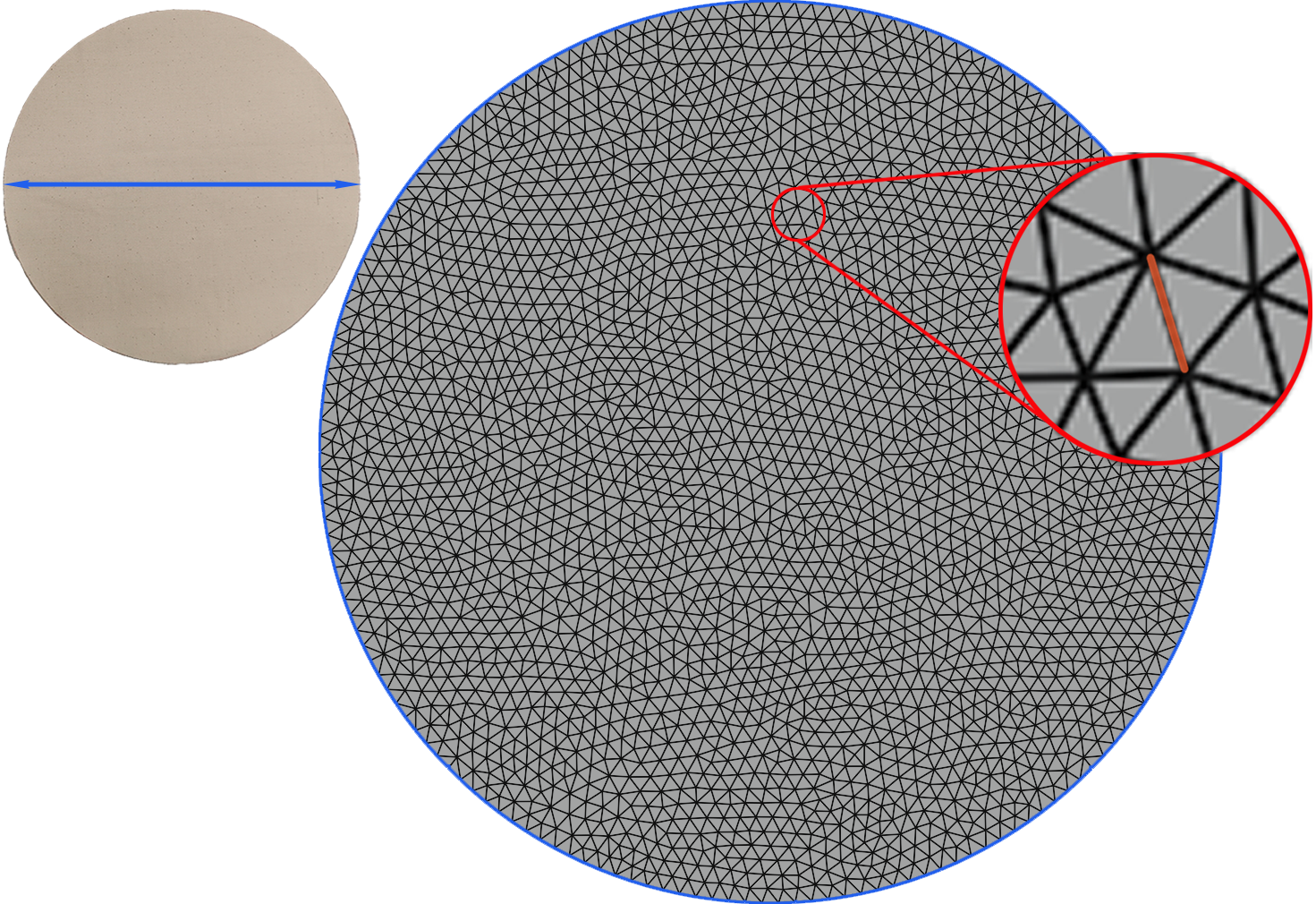}
        \put(10, 56){$30cm$}
        \put(10,36){(a)}
        \put(30,2){(b)}
    \end{overpic}   
    \caption{(a) A circular (diameter=$30cm$) cloth sample for Cusick drape test. (b) Cloth sample mesh has 2699 vertices, 7924 edges (7754 bending edges), and 5226 faces. A \textbf{bending edge} (highlighted in orange) is shared between two adjacent triangles, so the edges highlighted in blue (the boundary) are not bending edges.}
    \label{fig:Mesh}
\end{figure}

We discretize a cloth sample into a triangular mesh with $v$ vertices (\cref{fig:Mesh}(b)). Then we define the sample's state as $\mathcal{S}_t=\{\mathbf{x}_t, \dot{\mathbf{x}}_t\}$ where $\mathbf{x}_t \in \mathbb{R}^{3v}$ and $\dot{\mathbf{x}}_t \in \mathbb{R}^{3v}$ are the vertex position and velocity respectively at time $t$. Therefore, a draping motion with discretized time is $\mathcal{S}_{0:n} = \{\mathcal{S}_t : t \in \mathbb{Z}^{+}; t \leq n\}$, with a time step size $h$. Since we only observe the final image $\mathcal{I}$ and the initial state $\mathcal{S}_0$, their joint probability $p(\mathcal{I}, \mathcal{S}_0)=$
\begin{align}
    \label{eq:MarkovProb}
    \idotsint p(\mathcal{I} | \mathcal{S}_n, \tau)
    \prod_{i=0}^{n-1} p(\mathcal{S}_{i+1} | \mathcal{S}_i, \tau) p(\tau) d\mathcal{S}_{1:n} d\tau
\end{align}
where we introduce two sets of latent variables, $\tau$ and $\mathcal{S}_{1:n}$. $\tau$ is the physical parameters. $\mathcal{S}_{1:n}$ is the intermediate states of the draping motion which we cannot directly observe. Since the draping is a physical process, it is reasonable to assume $\mathcal{S}_t$ is only affected by $\mathcal{S}_{t-1}$ and $\tau$ (Markov assumption). Additionally, the captured image $\mathcal{I}$ is only decided by the final state $\mathcal{S}_n$.

\cref{eq:MarkovProb} is not easy to estimate due to two challenges. First, unlike the prior works which depend on dense observations on the intermediate state transitions \cite{yang2017learning} to estimate $p(\mathcal{S}_{i+1} | \mathcal{S}_i, \tau)$, Cusick drape testing does not capture the cloth sample's motion. Also, we do not observe the full $\mathcal{S}_n$, but only its (simplified) 2D representation $\mathcal{I}$. 
\begin{figure*}[tb]
    \centering
    \includegraphics[width=0.8\textwidth]{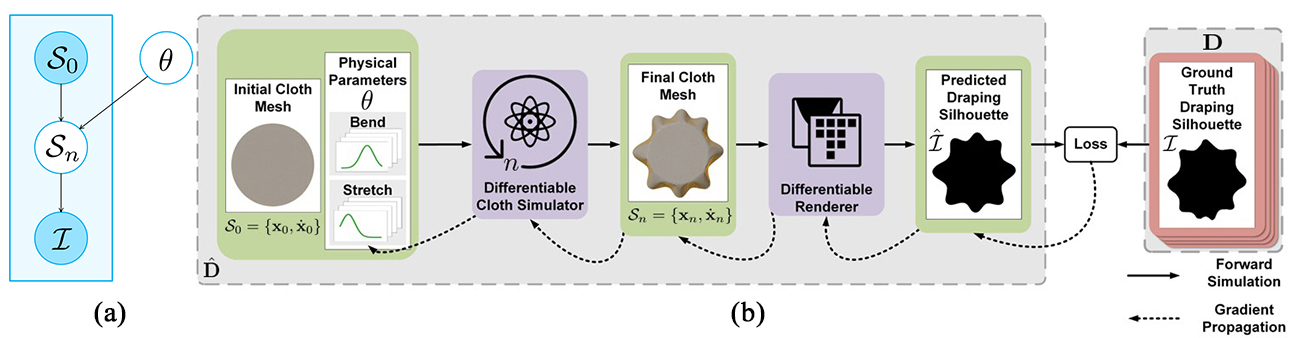}
    \caption{(a) The \textbf{Probabilistic Graphical Model} (PGM) of our Bayesian Differentiable Simulator. (b) \textbf{Model overview.} The physical parameters (stretching stiffness, bending stiffness) are first drawn from their learnable posteriors. Then the parameters and the cloth initial state are fed to a differentiable cloth simulator to run and predict cloth's final state $\mathcal{S}_n=\{\mathbf{x}_n, \mathbf{\dot{x}}_n\}$. The cloth in the final state is passed to a differentiable renderer. The rendered cloth silhouette is compared with the ground truth to compute the loss for back-propagation to update the parameters in the posteriors.}
    \label{fig:Model}
\end{figure*}

To this end, we assume two deterministic mappings can be established for $p(\mathcal{S}_{i+1} | \mathcal{S}_i, \tau)$ and $p(\mathcal{I} | \mathcal{S}_n, \tau)$. The determinism assumptions are reasonable as $p(\mathcal{S}_{i+1} | \mathcal{S}_i, \tau)$ can be seen as a quasi-deterministic physical process, subject to minor system stochasticity which is largely mitigated by the rigorous control of Cusick settings. $p(\mathcal{I} | \mathcal{S}_n, \tau)$ can be seen as a rendering process studied in computer graphics. Here we are mainly interested in the silhouette, following the practice in textile in Cusick drape analysis~\cite{chu1950mechanics}. Therefore, we replace $p(\mathcal{S}_{i+1} | \mathcal{S}_i, \tau)$ with a state transition function $\mathcal{S}_{i+1} = s(\mathcal{S}_i, \tau)$ where $s$ is a deterministic function then $p(\mathcal{S}_{i+1} | \mathcal{S}_i, \tau) = p(s(\mathcal{S}_i, \tau) | \mathcal{S}_i, \tau) = 1$. Similarly, $p(\mathcal{I} | \mathcal{S}_n, \tau) = p(r(\mathcal{S}_n) | \mathcal{S}_n, \tau) =  1$ where $r$ is a rendering function. Therefore, \cref{eq:MarkovProb} is transformed to:
\begin{equation}
\label{eq:prob_model}
    p(\mathcal{I}, \mathcal{S}_0) = \int p(r(\underbrace{s\dots s}_{n}(\mathcal{S}_0, \tau))) p(\tau )d\tau
\end{equation}
Given an initial state $\mathcal{S}_0$ and the observations 
$D=\{\mathcal{I}_1, \mathcal{I}_2,\dots \}$, we maximize $p(\mathcal{I}, \mathcal{S}_0)$ which is equivalent to finding the posterior distribution of cloth material parameters: $p(\tau|D) \propto p(D|\tau) p(\tau) = p(D | r \circ s \circ s \dots s(\mathcal{S}_0, \tau))p(\tau) = p(D|\hat{\mathcal{I}})p(\tau)$ where the composite function $\hat{\mathcal{I}} = r \circ s \circ s \dots s(\mathcal{S}_0, \tau)$ is deterministic. Overall, the corresponding Probabilistic Graphical Model (PGM) is illustrated in \cref{fig:Model} a.

\subsection{Model Specification}

\label{sec:modelSpec}
To infer $p(\tau| D)$, we need to instantiate $s$ and $r$. For $r$, we use differentiable renderer DIR-B \cite{chen2019learning}. Given a final draped state $\mathcal{S}_n$ which is a 3D mesh, and the virtual camera pose, DIR-B converts it to a 2D image. We set up the virtual camera pose (relative to the cloth drape) according to the real camera in our Cusick drape meter, so that the images captured in the Cusick drape test can be directly used as training data. Additionally, we only use drape silhouettes and ignore other information such as textures. This is because it is less reliable during capture and irrelevant to cloth drapability \cite{cusick196130,cusick196546,cusick196821}. 

\subsubsection{A Bayesian Differentiable Cloth Model}

The instantiation of $s$ is more complex than $r$. We propose to use a differentiable cloth model for $s$ so that we can use back-propagation for learning. Unlike the previous differentiable models, we also want to account for the material heterogeneity, and dynamics stochasticity. Therefore, we build a stochastic heterogeneous model, where we coin the term Bayesian differentiable cloth model. We give the key equations below and leave the details in the supplementary material (SM). By using implicit Euler method~\cite{baraff1998large}, the physical governing equation is defined as:
\begin{equation}
    \left( \mathbf{M} - h \frac{\partial \mathbf{f}}{\partial \dot{\mathbf{x}}} - h^2 \frac{\partial \mathbf{f}}{\partial \mathbf{x}} \right) \Delta \dot{\mathbf{x}} = h \left( \mathbf{F}_{t-1} + h \frac{\partial \mathbf{f}}{\partial \mathbf{x}} \dot{\mathbf{x}}_{t-1}\right)
\label{eq:motion}
\end{equation}
where $\mathbf{M}$ is the general mass matrix, function $\mathbf{f}$ takes as input the vertex position $\mathbf{x}$ and velocity $\dot{\mathbf{x}}$ to compute the resultant force $\mathbf{F}$: $\mathbf{F}_{t-1} = \mathbf{f}(\mathbf{x}_{t-1}, \dot{\mathbf{x}}_{t-1})$ where $\mathbf{F}_{t-1}$ is the resultant force at time $t-1$. The resultant force consists of all the internal and external forces: $\mathbf{F} = \mathbf{F}_{gravity} + \mathbf{F}_{handle} + \mathbf{F}_{stretch} + \mathbf{F}_{bend}$. $\mathbf{F}_{gravity}$ is simply gravity and $\mathbf{F}_{handle}$ is the force for pinning and supporting a cloth sample, \eg simulating the inner support panel (\cref{fig:testing}).

Deviating from previous methods~\cite{liang_2019_differentiable,Gong_finegrained_2022}, we model a material variation across the mesh which is discretized by finite elements. This allows us to localize the learning to each element, \ie. making the learning of $\mathbf{F}_{stretch}$ and $\mathbf{F}_{bend}$ dependent on local deformation. The stretching force~\cite{volino2009simple} on face $j$ is:
\begin{equation}
    \mathbf{F}_{stretch}^{(j)} = - A^{(j)}  
    \left(
        \sum_{m \in (uu, vv, uv)} \sigma_m^{(j)}
        \left(
            \frac{\partial \varepsilon_m^{(j)}}{\partial \mathbf{x}_i}
        \right)
    \right)
\end{equation}
where $A^{(j)}$ is the rest area of the mesh face $j$, $\varepsilon^{(j)}$ denotes the stretching strain, and the stretching stress $\sigma_m^{(j)} = \mathbf{C}^{(j)} \varepsilon^{(j)}_m$ where the stretching stiffness $\mathbf{C}^{(j)} \in \mathbb{R}^{6 \times 4}$. The subscripts $uu$, $vv$, and $uv$ denote strain/stress along cloth warp/wale, weft/course and diagonal direction respectively. Further, the bending force on a bending edge $w$ (\cref{fig:Mesh}) is defined as:
\begin{equation}
    \mathbf{F}_{bend}^{(w)} = \mathbf{B}^{(w)} \frac{|\mathbf{e}^{(w)}|}{\psi_1^{(w)} + \psi_2^{(w)}} \sin(\frac{\gamma^{(w)}}{2} - \frac{\bar{\gamma}^{(w)}}{2} )  u_i
    \label{eq:bending}
\end{equation}
where $\mathbf{B}^{(w)} \in \mathbb{R}^{3 \times 5}$ is the bending stiffness, $|\mathbf{e}^{(w)}|$ is the rest length of the bending edge, $\psi_1^{(w)}$ and $\psi_2^{(w)}$ are the height of the two triangular faces sharing the edge $w$. $\gamma^{(w)}$ and $\bar{\gamma}^{(w)}$ are the current and the rest dihedral angles between two faces \cite{tamstorf2013discrete}. Refer to \cite{bridson2005simulation} for the $u$'s. Overall, the learnable parameters $\tau =\{\mathbf{C}, \mathbf{B}\}$ is high dimensional (elaborated in the SM).

Now we replace the deterministic mapping $s$ in \cref{eq:prob_model} by solving \cref{eq:motion} forward in time, so that \cref{eq:prob_model} considers physical parameter variation. The prior $p(\tau)$ (Eq. \eqref{eq:prob_model}) is used as a belief of the parameter distribution and a posterior $p(\tau|D)$ is learned through inference.

\subsection{Model Inference}

Directly estimating $p(\tau|D)$ is computationally intractable. We adopt variational inference~\cite{hoffman2013stochastic} to seek an variational distribution $q_\theta(\tau)$ parameterized by $\theta$, to approximate the true posterior $p(\tau|D)$ by minimizing the Kullback-Leibler divergence between them:
\begin{align}
    \label{eq:VI}
    &\theta = \argmin_{\theta}D_{\mathbb{KL}}(q_{\theta}(\tau) \| p(\tau|D)) \notag \\
    &= \argmin_{\theta}\mathbb{E}_{q_{\theta}(\tau)} 
    \left[ 
    \log q_{\theta}(\tau) - 
    \log \left( \frac{p(D|\tau)p(\tau)}{p(D)}\right) 
    \right] \notag \\
    &= \argmin_{\theta}\mathbb{E}_{q_{\theta}(\tau)} [\log q_{\theta}(\tau) -\log p(D|\tau)p(\tau)] + \underbrace{\log p(D)}_{\mbox{const}} \notag \\
    &\equiv \argmin_{\theta}\underbrace{\mathbb{E}_{q_{\theta}(\tau)} [\log q_{\theta}(\tau) -\log p(D|\tau)p(\tau)]}_{\mathcal{L(\theta|D,\tau)}}
\end{align}
Calculating $\mathcal{L(\theta|D,\tau)}$, the negative evidence lower bound, is computationally prohibitive. So we approximate it by Monte Carlo sampling:
\begin{equation}
    \label{eq:MC}
    \mathcal{L(\theta|D,\tau)} \approx \sum_{i=1}^m
    \big(
    \underbrace{\log q_{\theta}(\tau_i) - \log p(D|\tau_i)p(\tau_i)}_{l(\theta,\tau)}
    \big)
\end{equation}
where $\tau_i$ denotes the \textit{i}th Monte Carlo sample from the variational posterior distribution $q_{\theta}(\tau)$. Moreover, we assume that the cloth physical parameters are distributed as a Gaussian $\tau \sim \mathcal{N}(\mu, \Sigma)$ where $\Sigma$ is a diagonal matrix. To enable stochastic gradient back-propagation, we adopt the re-parameterization trick~\cite{blundell2015weight, kingma2013auto} to sample physical parameters $\tau = t(\epsilon, \theta)$ by shifting a stochastic parameter-free noise $\epsilon \sim \mathcal{N}(0, \mathbf{I})$ through the deterministic function $t(\epsilon, \theta) = \mu + \log{(1 + exp(\eta))} \odot \epsilon$ where variational parameters $\theta=\{\mu, \eta \}$. Consequently, the variational distribution $q_{\theta}$ is sought within the Gaussian family and the prior $p(\tau)$ is an isotropic Gaussian distribution with fixed parameters. Additionally, the output distribution is also a Gaussian $\mathcal{N}(\mu_I, \sigma^2)$ whose mean depends on predicted image $\hat{\mathcal{I}}$, i.e. $\mu_I=\hat{\mathcal{I}}$. The variance, $\sigma^2$, is fixed and used to control the tolerance to residual error. Therefore, the negative log likelihood $-\log p(D|\tau)$ is:
\begin{equation}
     - \sum_{i=1}^{L}\sum_{j=1}^{L}\log 
    \left[
    \left(\frac{1}{2\pi\sigma^2}\right)^{\frac{1}{2}}
    e^{ -\frac{1}{2\sigma^2}(\mathcal{I}_{ij} - \hat{\mathcal{I}}_{ij})^2}
    \right]
\end{equation}
which is essentially proportional to the Mean Squared Error (MSE). In back-propagation, the gradients of the variational distribution parameters are calculated by\cite{blundell2015weight}:
\begin{gather}
    \frac{\partial \l(\theta, \tau)}{\partial \mu} = \frac{\partial l(\theta, \tau)}{\partial \tau} + \frac{\partial l(\theta, \tau)}{\partial \mu} \\
    \frac{\partial \l(\theta, \tau)}{\partial \eta} = \frac{\partial l(\theta, \tau)}{\partial \tau} \frac{\epsilon}{1 + e^{-\eta}} + \frac{\partial l(\theta, \tau)}{\partial \eta}
\end{gather}
The training/inference algorithms are detailed in the SM.

\subsection{Implementation}

Our differentiable cloth simulator is implemented in Pytorch's C++ frontend \cite{paszke2019pytorch}. We exploit vectorization and CUDA GPU parallel computing for fast simulation and learning (in the SM). We use Eigen's sparse solver \cite{eigenweb} to solve the governing equations \cref{eq:motion}. To reduce the memory consumption, we use sparse matrices whenever possible. Moreover, we do in-place gradient update for every time step for back-propagation, so our memory usage does not increase with simulation steps \cite{chen2016training}. We use the Kaolin differentiable rendering package \cite{jatavallabhula2019kaolin} for image rendering. 

\section{Data Collection} 

The Cusick drape data is collected following the BS EN ISO 9073-9:2008~\cite{cdi_bsi_primary_000000000030128826}. In every test, the warp/wale and weft/course directions are aligned across samples to ensure the same initial condition. In addition, we are able to reconstruct the 3D meshes of cloth drapes thanks to our patented drape meter. However, the 3D data are mainly for evaluation and the silhouette images are for learning. This is because not every Cusick drape meter can reconstruct 3D meshes out of cloth drapes. Being able to learn from the 2D silhouette images only is crucial in making our method applicable in real-world settings. In addition, we also measure the sample weight and calculate the average area density $\rho = (\sum_{i<13}^{i=1} m_{i}) / (12 \pi R^2) $ where $m_i$ denotes the measured weight of cloth sample $i$ and $R=0.15m$. We also measure the thickness and include them in our dataset (refer to SM).

\section{Experiments}

Due to the nature of limited data, we conduct our experiments on small datasets. We use 5 representative types of cloths (each with 12 samples) that exhibit visually distinguishable drapability. For convenience, we name cloths as ``material color'', e.g. Cotton Blue, Cotton Pink. To digitalize a cloth, we randomly select 1 out of 12 samples for training to learn their parameter distributions. During learning, we run 100 steps ($n=100$) for the forward simulation, with time step size $h=0.05s$ and the total time lapse is $5s$. To qualitatively evaluate the digitalization, we simulate garments made from different digitalized cloths and visually compare the simulated garments with their corresponding real cloths. To evaluate the model (i.e. drape fitting), we employ several metrics. For fitting capability, we use mean squared error (MSE) between the fitted and the ground truth drape images. We also use Hausdorff distance (H.Dis) between the simulated 3D mesh and the ground-truth (GT) mesh. Further, we use radius-angle graph (\cref{fig:drape_radius})~\cite{kim2011determination} which is widely adopted for describing Cusick drape waves and comparing drape shapes. As generalization, we also test if our model learned on one sample can predict the physics of unseen cloth samples of the same type. \url{https://youtu.be/ProN0y1bURY} has more visual results.

\subsection{Cloth Digitalization for Garments}

Our model can digitalize real cloths for garment simulation, by sampling from the learned parameter distributions. We show three representative cloths (\cref{fig:teaser} (a-1, b-1, c-1)) with distinctive silhouette and 3D drape shapes due to their diversified drapability. Given a skirt geometry, we sample from the learned parameter distributions for the stretching and bending stiffness for each mesh triangle for simulation, shown in \cref{fig:teaser} (a-2, b-2, c-2). Within each cloth, both static (left) and dynamic (right) drapes are shown. Visually, the drapability of the skirts are similar to their corresponding cloths above. For instance, the Viscose White has a small drape shape because it is heavy and soft (small bending stiffness). By contrast, the Wool Red has a much larger drape shape because it is thick and stiff (large bending stiffness). Correspondingly, the simulated Wool Red skirt looks wider in the static and has larger folds in motions. Here, being able to learn the possible distributions of physical parameters enables us to apply the learned material to garments with arbitrary geometries.

\begin{figure}[tb]
    \centering
    \includegraphics[width=0.47\textwidth]{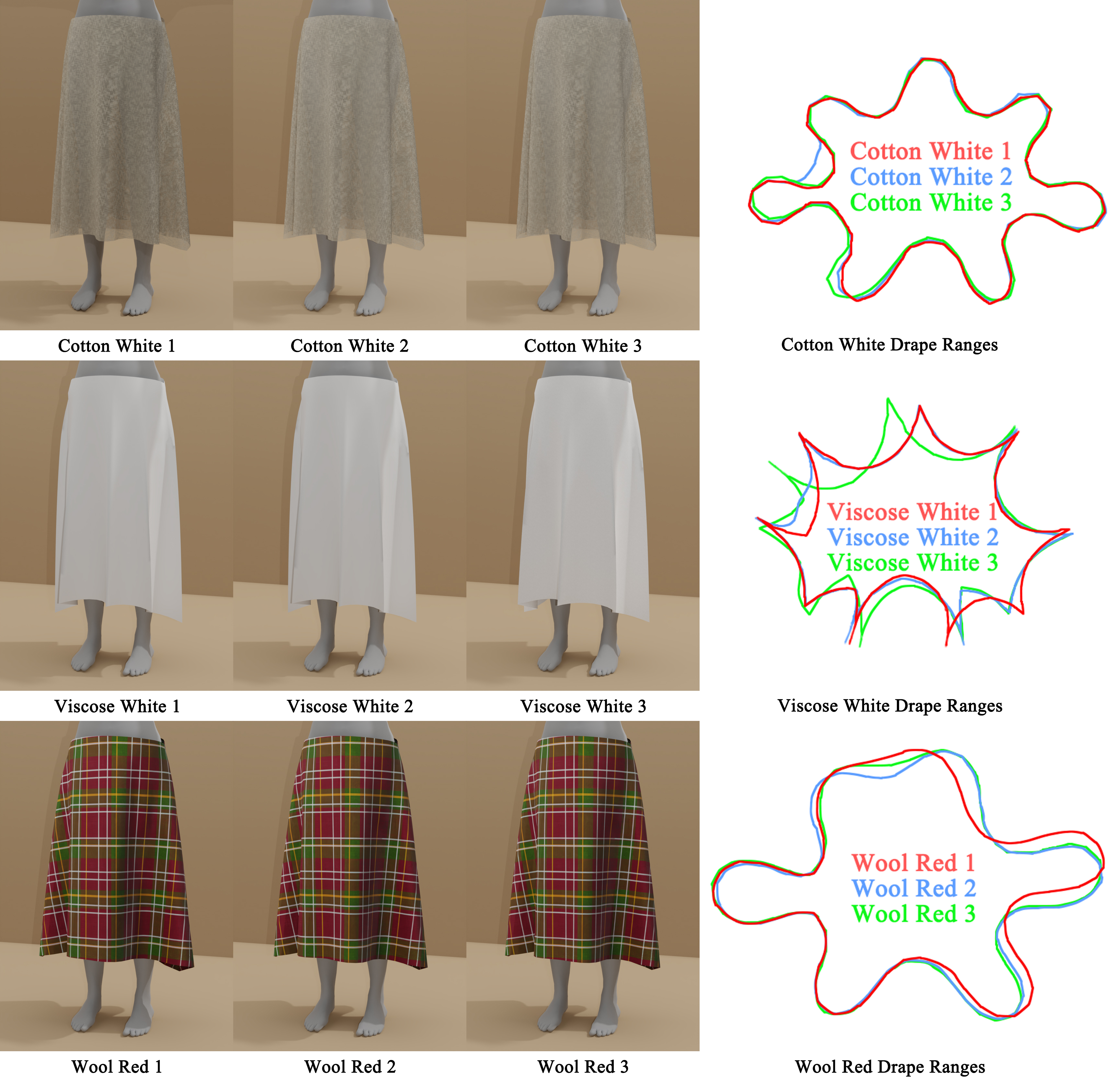}
    \caption{Given the digitialized cloths, our BDP model can simulate the skirts made from these cloths and reflect cloth material heterogeneity and draping stochasticity.}
    \label{fig:bnn_dress}
\end{figure}

Apart from cross-material differences, our Bayesian Differentiable Physics (BDP) can also capture the material heterogeneity and dynamics stochasticity within the same cloth type. \cref{fig:bnn_dress} left shows the simulated skirts made from the same three cloths as in \cref{fig:teaser}. Three sets of parameters are sampled for each cloth type and we also show the silhouette of the drape of the skirt in \cref{fig:bnn_dress} right. Within each cloth type, the material heterogeneity and dynamics stochasticity can be observed in both the 3D drapes and the 2D silhouette, while the overall distributions of them across types are significantly different each other. This shows BDP captures both the cross-material and within-type heterogeneity, and dynamics stochasticity well. 

\subsection{Comparison}

\paragraph{Alternative Cloth Models} 

To our best knowledge, there is no similar method designed for exactly the same setting as ours. We therefore adopt~\cite{liang_2019_differentiable} the closest methods as baselines because their method can also digitalize cloth by learning cloth physical parameters, albeit only taking simulation data as input. Since they only model homogeneous material, we refer to their model as HOMO. Further, we augment their model by making the cloth physical parameter learning element-wise, to enable learning heterogeneous materials. We call this model HETER for short. Both HOMO and HETER are \textit{deterministic} models. So we use MSE as the loss function: $\mathcal{L}(\tau) = \sum_{i=1}^{L}\sum_{j=1}^{L}(\hat{\mathcal{I}}_{ij} - \mathcal{I}_{ij})^2$, for training. For comparison, we train HOMO and HETER with the same data (1 for training and the rest 11 for testing) as BDP. For BDP, we draw parameters from the learned parameter distributions for 1000 times and run Cusick drape simulations. Then, we select the best result to calculate the MSE and H. Dis. 

\begin{table}[tb]
    \centering
    \begin{tabular}{cccc}
        \toprule
        Metrics &  HOMO & HETER & BDP   \\
        \midrule
        Avg MSE &  \num{5.72e-2} & \num{5.51e-2} & \num{3.84e-2} \\
        Avg H. Dis & \num{3.40e-2} & \num{3.28e-2} & \num{2.17e-2} \\
        \bottomrule
    \end{tabular}
    \caption{MSE and H. Dis (meter) of HOMO, HETER and BDP.}
    \label{tab:comp_all}
\end{table}

\begin{figure}[tb]
    \centering
    \includegraphics[width=0.47\textwidth]{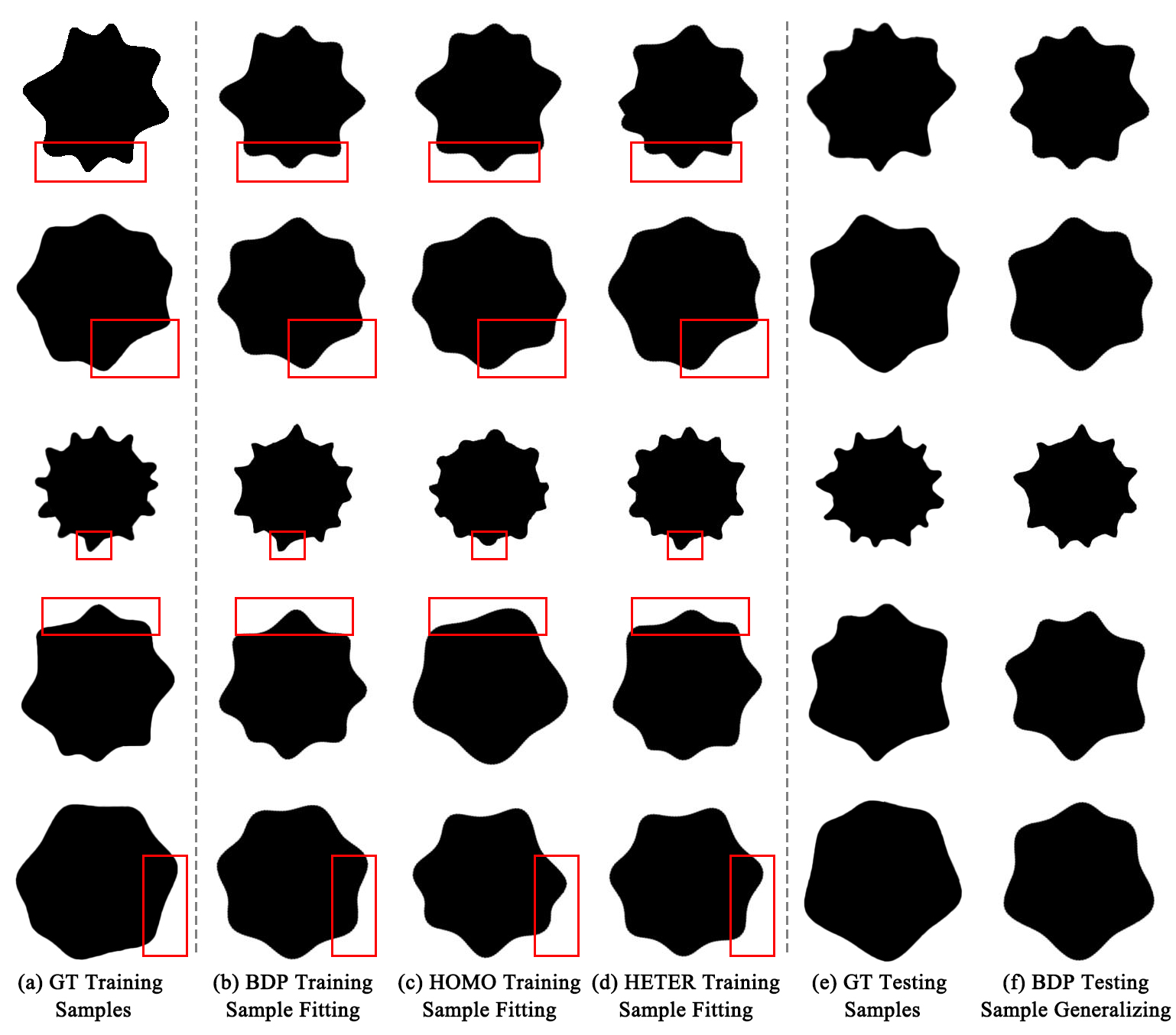}
    \caption{(a), (b), (e), and (f) show that BDP can fit the training sample and generalize to the unseen testing samples. By contrast, the learned drape shapes by HOMO (c) and HETER (d) are different from the testing sample. As such, only BDP can capture/reproduce cloth within-type variations and is superior in cloth digitalization. Additionally, by modeling material heterogeneity, HETER fit the training sample more accurately than HOMO.}
    \label{fig:compare_all}
\end{figure}
\begin{figure}[htb]
    \centering
    \includegraphics[width=0.47\textwidth]{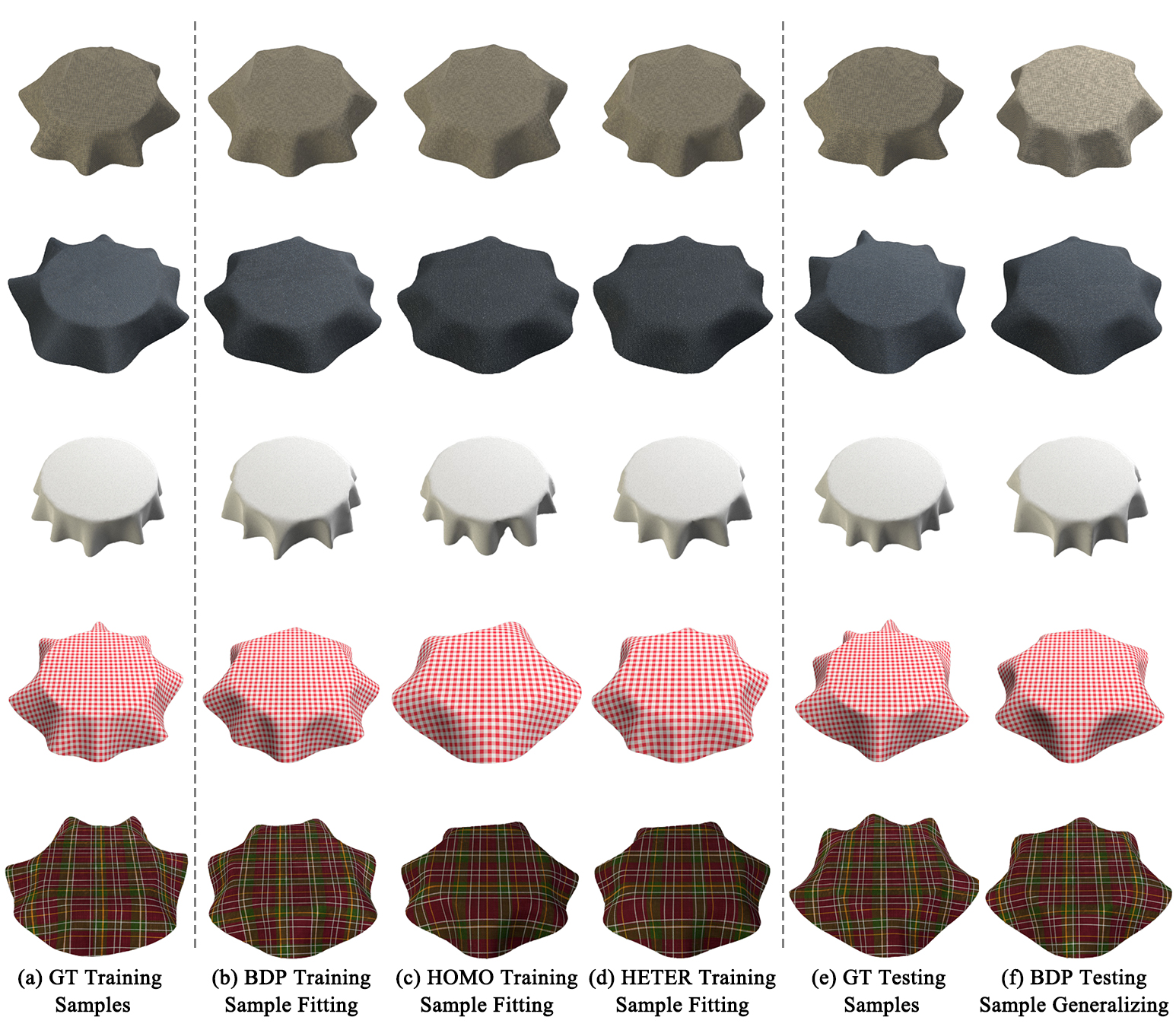}
    \caption{The rendered drape shapes in \cref{fig:compare_all}.}
    \label{fig:compare_all_render}
\end{figure}

As shown in \cref{tab:comp_all} (and the corresponding rendered cloths in \cref{fig:compare_all_render}), our BDP outperforms HOMO and HETER. Visually, our method not only can accurately fit the training sample (\cref{fig:compare_all} (b)), but can also predict the testing sample (\cref{fig:compare_all} (f)), demonstrating that our model can capture cloth within-type variations. Unsurprisingly, HOMO and HETER are deterministic models and can only learn from and reproduce the training sample. Furthermore, compared with HOMO, HETER fits the training sample better and this demonstrates the importance of modeling cloth material heterogeneity. However, HETER alone is inadequate for cloth digitialization because the estimated physical parameters are tied to the sample geometry and it is difficult to generalize it to garments. Although HOMO can theoretically generalize to garments, it can only simulate the material specific to the training sample, not being able to generalize to similar materials within the same cloth type.

\begin{figure}[tb]
    \centering
    \subcaptionbox{An example of draping radius \label{fig:drape_radius}}{
    \includegraphics[width=0.47\textwidth]{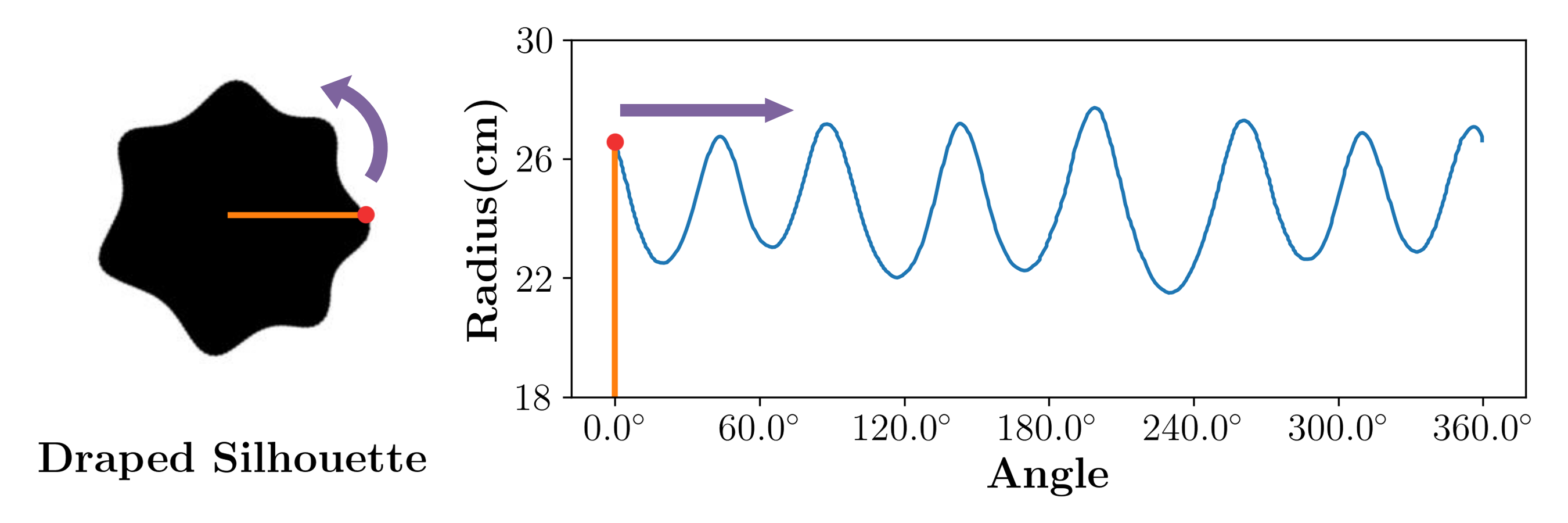}}
    \\
    \subcaptionbox{Clustered drape radius \label{fig:cluster}}{
    \includegraphics[width=0.47\textwidth]{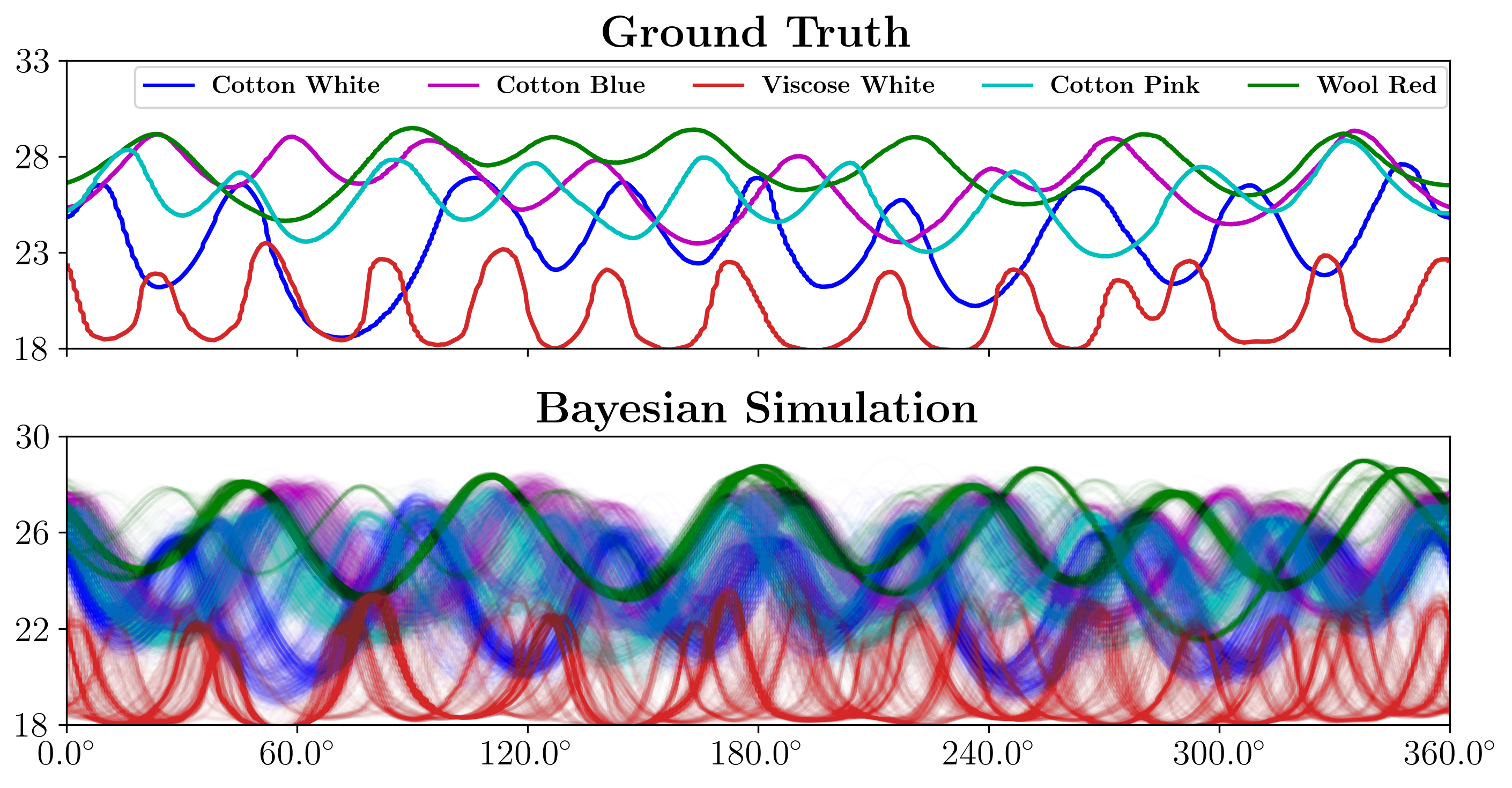}}
    \caption{\subref{fig:drape_radius}: an drape radius-angle graph illustrates the varied radius of an draped cloth sample's boundary w.r.t. angle~\cite{kim2011determination}. \subref{fig:cluster}: the ground truth from five real cloth samples (top) and the simulated 1000 samples from our BDP (bottom) for the corresponding samples. It demonstrates our trained BDP models can distinguish different cloths and are not overly generalized.}
    \label{fig:exp3_cluster}
\end{figure}
Although BDP can learn within-type material variations (\cref{fig:compare_all}), one question is whether it overly generalizes, \ie unable to distinguish different cloth types. To this end, we use radius-variation~\cite{kim2011determination} to demonstrate that BDP does not overly generalize. In \cref{fig:cluster}, the top figure shows that the five real cloth samples have distinctive drape shapes. The bottom figure shows the drape shapes of samples from our BDP which largely follow the same patterns as the ground truth. For example, Wool Red is obviously stiffer than Viscose White, as shown in the Ground Truth. Likewise, it is also observed in our BDP's simulation at the bottom. Refer to the SM for more results.

\paragraph{Alternative Learning Methods}

\begin{table}[bt]
    \centering
    \begin{tabular}{cccc}
        \toprule
        Metrics  & BDP & REMBO & HeSBO \\
        \midrule
        Avg MSE  & \num{4.52e-2} & \num{6.03e-2} & \num{5.90e-2}\\
        Avg H. Dis & \num{2.01e-2} & \num{4.25e-2} & \num{4.35e-2}\\
        \bottomrule
    \end{tabular}
    \caption{The average MSE and H. Dis of ours, REMBO, and HeSBO optimizer. The results shows that our gradient-based optimization achieves better results.}
    \label{tab:comp_opt}
\end{table}
While our method is built on differentiable physics models for learning, \ie derivative-based optimization, the traditional methods widely used in material science and physics are usually based on derivative-free optimization. So we also compare different learning strategies. Bayesian Optimization (BO) is a representative derivative-free optimizer which usually uses the Gaussian Process to approximate an unknown optimized objective function \cite{brochu2010tutorial, frazier2018tutorial}. However, the performance of vanilla BO drops drastically when the number of parameters is above 20~\cite{eriksson2021high, letham2020re}. There are over 200,000 parameters in the HETER cloth model. So we employ Random Embedding Bayesian Optimization (REMBO)~\cite{wang2016bayesian} and Hashing-enhanced Subspace Bayesian Optimization (HeSBO)~\cite{nayebi2019framework} as baselines. We use BoTorch's \cite{balandat2020botorch} REMBO and HeSBO implementation and compare them with our BDP. Given the same drape silhouette, we run our method 1000 epochs (gain the best optimization result within 200 epochs), and the REMOB and HeSBO for 500 trials. \cref{tab:comp_opt} shows that, our derivative-based method is better with fewer optimization steps. More results can be found in the SM.

\section{Conclusion and Future Work}

We have proposed a new method for cloth digitalization by estimating detailed cloths physical properties. To our best knowledge, this is the first Bayesian differentiable cloth model that can work seamlessly with standard Cusick drape data. Our model has been proven to be highly accurate and generalizable. Despite focused on cloth digitalization, we believe BDP as a framework has the potential to generalize to more generic digitalization tasks and itself is a methodological extension of current DP research in computer vision. Compared with black-box deep learning methods, our limitation is that it requires prior knowledge of the underlying physics and cannot simply \textit{plug and play} on data. However, we argue that this is a reasonable trade-off when data collection is expensive and slow. In future, we plan to model more dynamics stochasticity, \eg buckling, and also compare our simulated garments with real ones by using accurate motion capture and 3D reconstruction systems.

\section*{Acknowledgements}
The project is partially supported by the Art and Humanities Research Council (AHRC) UK under project FFF (AH/S002812/1).


{
    \small
    \bibliographystyle{ieeenat_fullname}
    \bibliography{main}
}

\end{document}







\clearpage
\setcounter{page}{1}
\maketitlesupplementary


\section{Cusick Drape Dataset}

\begin{table*}[tb]
    \centering
    \begin{tabular}{cccccc}
        \toprule
        Fabric Index & Material & Woven & \# Samples & Avg $\rho$ ($kg/m^2$) & Avg Thickness($mm$) \\
        \midrule
        Fabric 1 & Cotton & Plain & 12 & 0.059 & 0.188 \\
        Fabric 2 & Cotton & Basket & 12 & 0.192 & 0.402 \\
        Fabric 3 & Viscose(95\%) Elastane (5\%) & Knit & 12 & 0.213 & 0.560 \\
        Fabric 4 & Cotton & Plain & 12 & 0.114 & 0.200 \\
        Fabric 5 & Wool & Twill & 12 & 0.274 & 0.571 \\
        Fabric 6 & Polyester & Satin & 12 & 0.183 & 0.240 \\
        Fabric 7 & Polyester(65\%) Cotton(35\%) & Plain & 12 & 0.100 & 0.195 \\
        Fabric 8 & Linen & Plain & 12 & 0.230 & 0.485 \\
        Fabric 9 & Cotton & Plain & 12 & 0.249 & 0.423 \\
        Fabric 10 & Viscose (70\%) Polyester (30\%) & Plain & 12 & 0.204 & 0.498 \\
        Fabric 11 & Wool & Twill & 2 & 0.148 & 0.210 \\
        Fabric 12 & Cotton & Plain & 2 & 0.313 & 0.188 \\
        Fabric 13 & Cotton & Plain & 2 & 0.107 & 0.15 \\
        Fabric 14 & Cotton & Plain & 2 & 0.139 & 0.292 \\
        Fabric 15 & Cotton & Plain & 2 & 0.211 & 0.436 \\
        Fabric 16 & Synthetic & Knit & 2 & 0.191 & 0.514 \\
        Fabric 17 & Cotton & Twill & 2 & 0.278 & 0.7 \\
        Fabric 18 & Synthetic & Plain/Knit & 2 & 0.154 & 0.306 \\
        Fabric 19 & Synthetic & Plain/Knit & 2 & 0.282 & 0.627 \\
        Fabric 20 & Synthetic & Twill & 2 & 0.259 & 0.596 \\
        Fabric 21 & Synthetic & Knit & 5 & 0.186 & 0.7 \\
        Fabric 22 & Synthetic & Plain & 5 & 0.231 & 0.386 \\
        Fabric 23 & Cotton & Knit & 5 & 0.163 & 0.704 \\
        Fabric 24 & Cotton & Plain & 5 & 0.104 & 0.200 \\
        Fabric 25 & Cotton & Plain & 5 & 0.060 & 0.152 \\
        \bottomrule
    \end{tabular}
    \caption{Fabric information in our Cusick drape dataset. Fabric 18 and Fabric 19 are double-layer fabrics whose one side is plain woven and the other side is knit.}
    \label{tab:drape_data}
\end{table*}

Our current Cusick drape dataset includes 25 types of common fabrics, each of which with multiple samples. As listed in \cref{tab:drape_data}, these fabrics are different in material, woven pattern, area density ($\rho$), and average thickness. In our Cusick drape test, they also show distinctive drape shapes. We test each sample multiple times. In every test, our Cusick drape meter captures a drape image and reconstructs its 3D mesh. It took approximately more than 15 working days to do the textile testing. At the end, there are 660 drape images and meshes in our current dataset which will be released with our paper. Fabric 1-5 are used in our experiments which are the Cotton White, Cotton Blue, Viscose White, Cotton Pink, and Wool Red respectively. 

The lack of training data is the deciding factor that hinders the application of deep learning in textile. To our best knowledge, the only open source cloth drape dataset was proposed in~\cite{feng2022learning} very recently. Compared with their dataset, our new dataset has multiples advantages. First, our dataset includes 25 types of common fabrics and has more training samples, which comprehensively covers a wide range of fabrics, with accurate description of the materials. The number of types of fabrics and details are unclear in their data.  Second, their data only provides the estimated material parameters tied to their estimation methods, which might make it difficult to transfer them to a different cloth model. In contrast, not only do we provide the raw Cusick drape testing data (\ie images and meshes), we also provide the estimated parameters. Furthermore, their parameters have specific values while ours are distributions that capture cloth material heterogeneity and dynamics stochasticity. Last but not the least, our Cusick drape testing rigidly follows the British standards and the textile measurements (e.g. measuring area density and thickness) are conducted in a rigorously controlled environment so that they are more accurate and easier to be reproduced/verified in future research. By contrast, their drape testing is measured by customized apparatus under a less controlled setting.

\section{Performance}

\begin{table}[tb]
    \centering
    \begin{tabular}{lccc}
    \toprule
    Test (sec/step) &  Mesh 1 & Mesh 2 & Mesh 3 \\
    \midrule
    \cite{liang_2019_differentiable} Forward & 0.643 & 2.793 & 3.861 \\
    \cite{liang_2019_differentiable} Backward & 1.400 & 7.379 & 26.212 \\
    Ours Forward & 0.038 & 0.073 & 0.178 \\
    Ours Backward & 0.062 & 0.228 & 0.660 \\
    \bottomrule
    \end{tabular}
    \caption{Comparing the run time between \cite{liang_2019_differentiable} (collision handling off) and our model in forward simulation and backward gradient computation. Our implementation is much faster.}
    \label{tab:comp_eff}
\end{table}

Simulation efficiency is critical to our cloth digitalization method because (1) training BDP usually requires longer time than deterministic models, and (2) generalizing the digitialized cloths on garments needs to simulate larger and more complex meshes. Our implementation of the differentiable cloth model is highly inspired by \cite{liang_2019_differentiable}. But ours is more GPU-friendly, and therefore runs fast. We compare the forward simulation and the backward gradient computation, which are two major time-consuming operations for differentiable physics models, between our implementation and \cite{liang_2019_differentiable}. We use three meshes with different resolutions, consisting of 279, 1205, and 2699 vertices respectively. \cref{tab:comp_eff} shows the significant performance gain by our vectorization and GPU parallel computing. The test is conducted on a PC with an Intel Xeon E5-1650 v4 3.60GHz CPU and an NVIDIA TITAN Xp GPU.

\section{Additional Results}

\subsection{Necessity of Material Heterogeneity} 
\begin{figure}[htb]
    \centering
    \includegraphics[width=.47\textwidth]{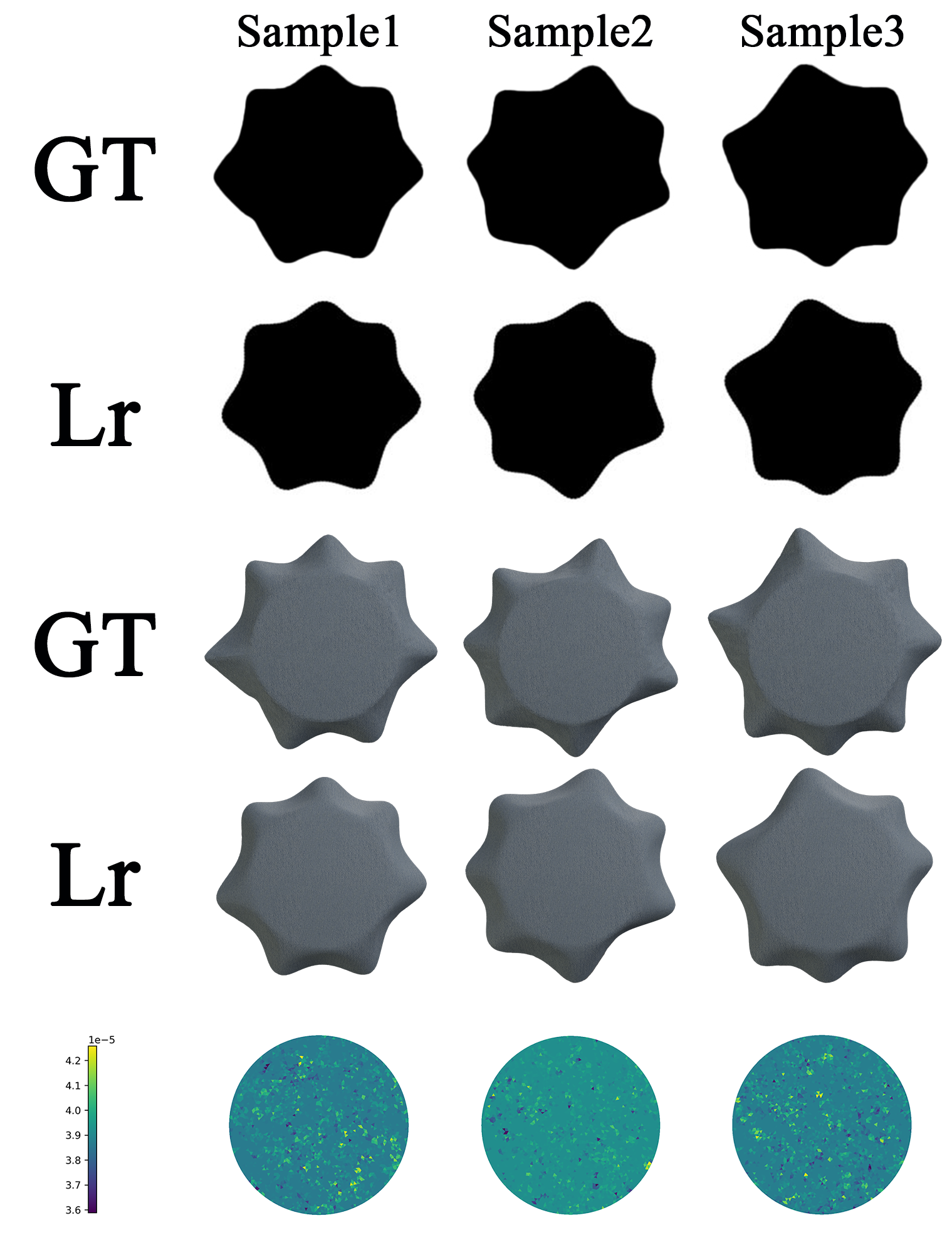}   
    \caption{Learning from drape shapes of the 3 samples cut from fabric Cotton Blue. The GT rows show the within-sample variations in drape shapes. The LR rows show the HETER model accurately learns these different drape shapes. The heat maps (bottom) show distributions the per-element bending stiffness $\mathbf{B}$ over the mesh.}
    \label{fig:cotton_blue_3_samples}
\end{figure}

\begin{figure}[htb]
    \centering
    \includegraphics[width=.47\textwidth]{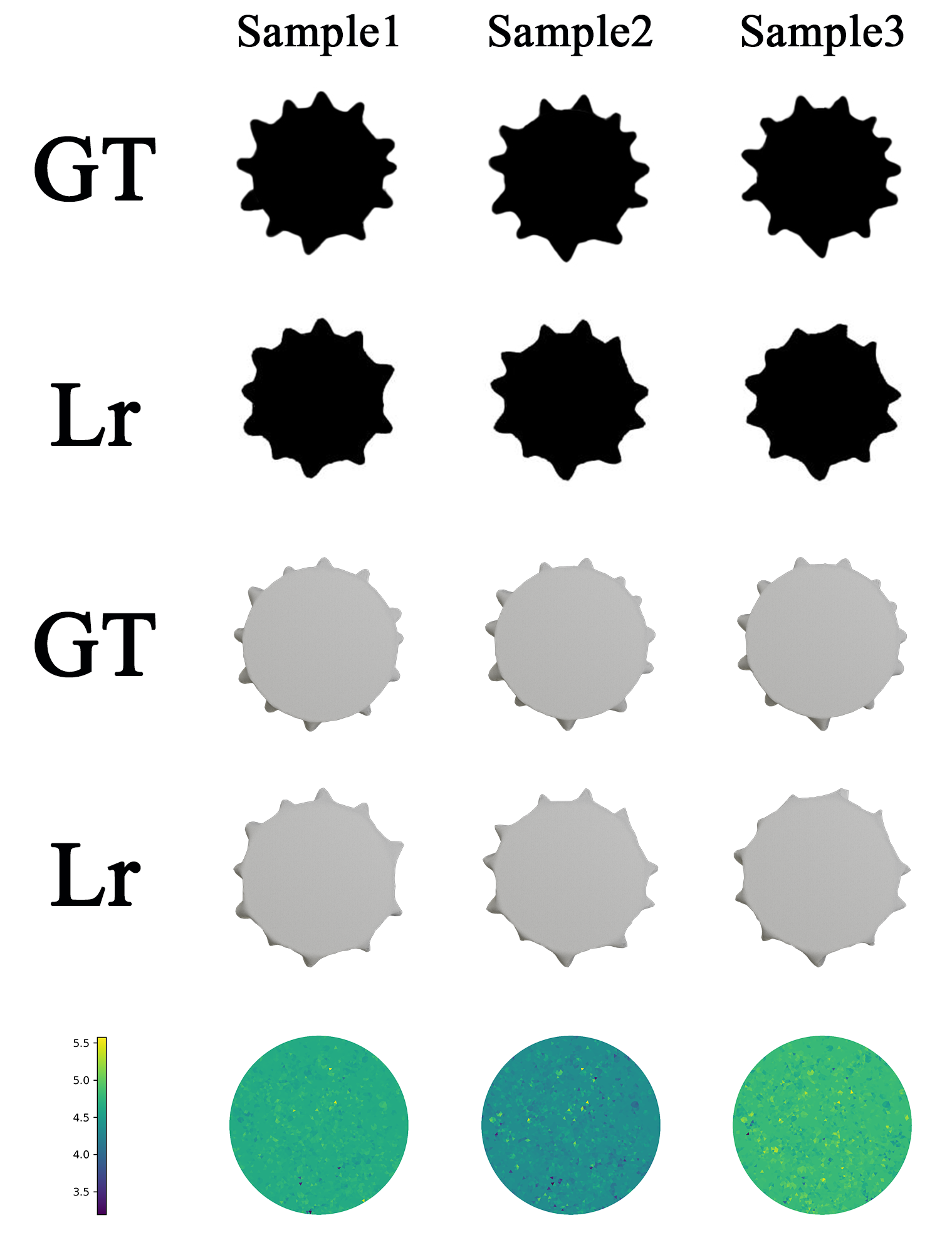}
    \caption{Learning from drape shapes of the 3 samples cut from fabric Viscose White. The GT rows show the within-sample varied draped shapes. The LR rows show the HETER model accurately learns these different drape shapes. The heat maps (bottom) show distributions the per-element stretching stiffness $\mathbf{C}$ over the mesh.}
    \label{fig:vicose_white_3_samples}
\end{figure}

Our model attributes the within-sample shape variation to cloth material heterogeneity. To demonstrate it, we use the HETER (deterministic heterogeneous model) to learn the drape shapes of six samples (three from Cotton Blue and three from Viscose White). The ground-truth (GT) rows in \cref{fig:cotton_blue_3_samples} and \cref{fig:vicose_white_3_samples} show that samples from the same fabric are obviously different. The learned (LR) rows in \cref{fig:cotton_blue_3_samples} and \cref{fig:vicose_white_3_samples} demonstrate that the HETER can accurately learn from the different drape shapes. The heat maps in the last row illustrate the learned distributions of the cloth physical parameters across the meshes. This result demonstrates that within-sample drape shape variation can be accounted for by cloth material heterogeneity and varied physical parameter distributed across the mesh in different samples.

\subsection{Learning from Meshes}

\begin{figure}[htb]
    \centering
    \includegraphics[width=.47\textwidth]{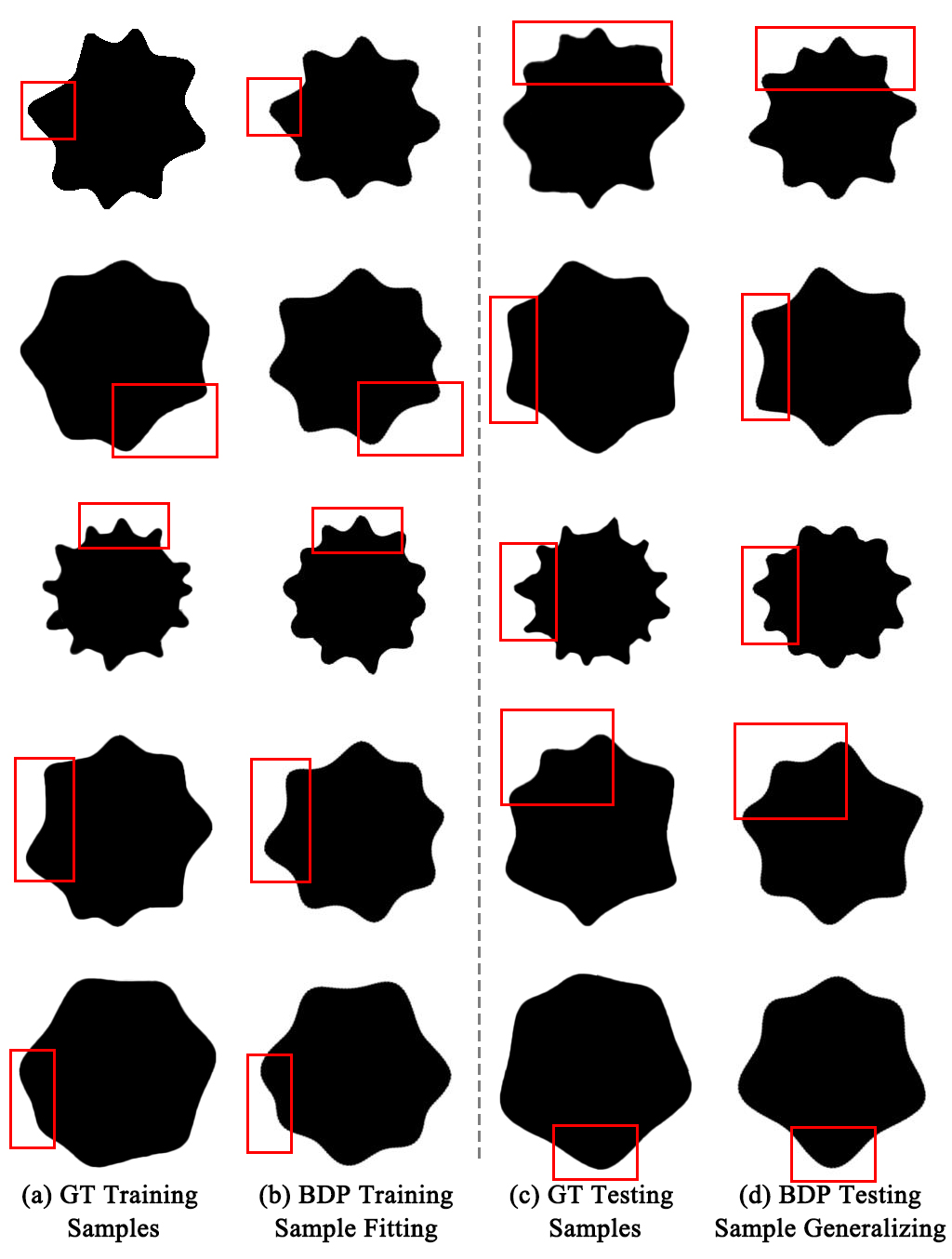}
    \caption{When learning from meshes, BDP can also fit the training sample(a) and generalize to the unseen testing samples(c).}
    \label{fig:abblation_mesh}
\end{figure}

We train the model using the reconstructed 3D meshes as the ground-truth which include geometry information, by minimizing the MSE between the simulated and real meshes. As shown in \cref{fig:abblation_mesh}, our BDP also exhibits outstanding training sample fitting and generalization ability. Quantitatively, the MSE and H.Dis are 0.051 and 0.023 under the same setting as Table 1 in the paper. Comparatively, 3D meshes do not improve the results and most testers cannot generate 3D mesh data.

\subsection{Effectiveness of Learning} 

\begin{table*}[htb]
    \centering
    \begin{tabular}{lccccc}
    \toprule
         Cloths &  Cotton White & Cotton Blue & Viscose White & Cotton Pink & Wool Red \\
         \midrule
         Avg. Stretching & 56.6019 & 63.9698 & 4.5117 & 101.4625 & 78.3332 \\
         Avg. Bending & \num{1.008e-05} & \num{8.249e-05} & \num{4.991e-05} & \num{4.829e-05} & 0.0001  \\
         Avg. Std & \num{1.435e-06} & 0.0001 & 0.0214 & \num{3.860e-06} & \num{2.057e-05} \\
         \bottomrule
    \end{tabular}
    \caption{The average stretching and bending stiffness of the five select cloths reflect their distinguishable physical properties. In addition, their average physical parameter standard deviations show their different material heterogeneity and dynamics stochasticity.}
    \label{tab:cloth_parameters}
\end{table*}

\begin{figure*}[tb]
    \centering
    \includegraphics[width=\textwidth]{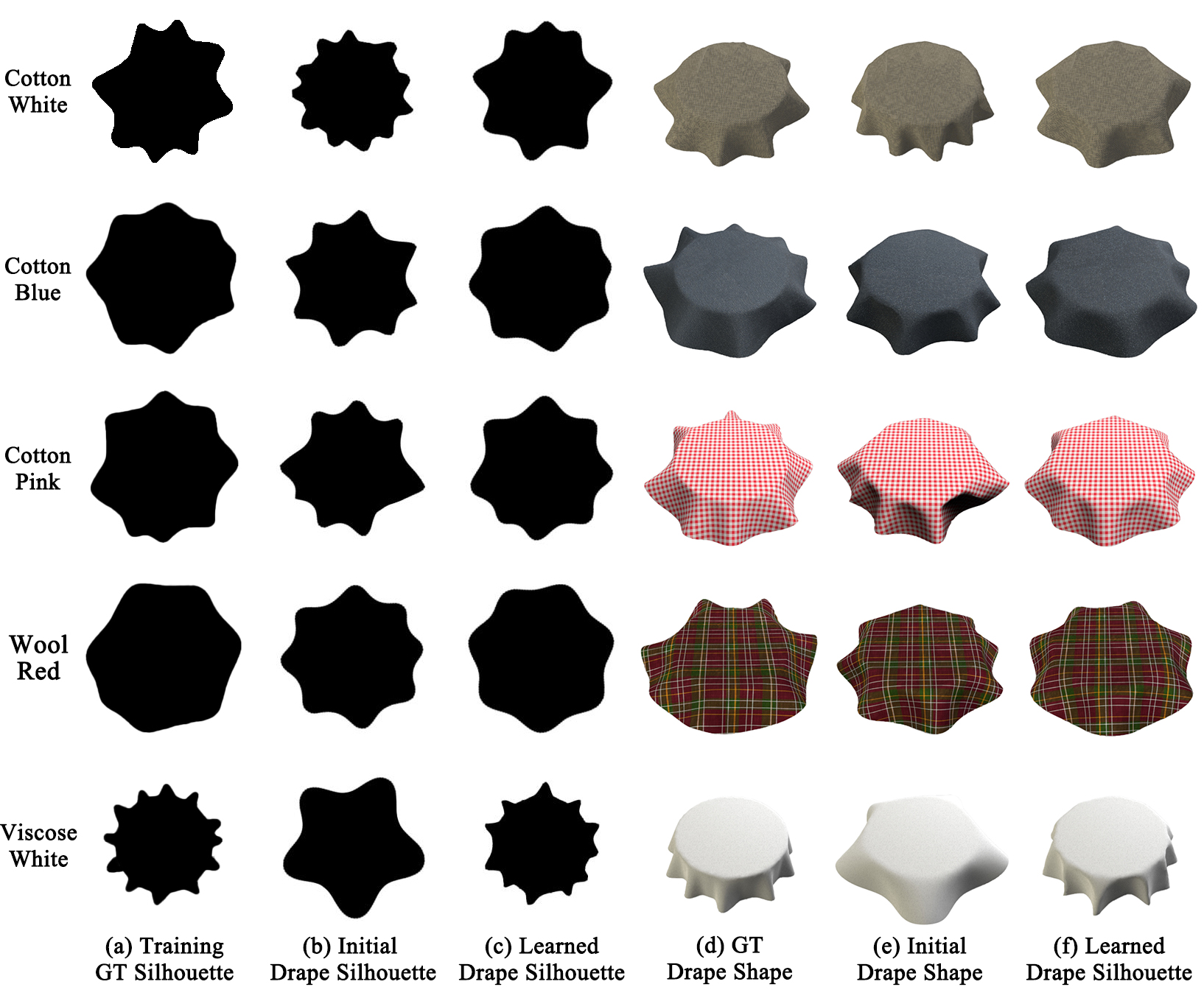}
    \caption{Given the initial parameters, the simulated drape shapes (b and e) are different from the training samples. After training, our BDP optimize cloth physical parameters and fit the given GT drape silhouettes (c and f).}
    \label{fig:bnn_train}
\end{figure*}

To digitalize cloths, our BDP uses gradient-based optimization to find cloth physical parameters such that cloth drape shape can be optimized toward the given ground truth sample. \cref{fig:bnn_train} shows that the optimization is effective, where the initial guess on the parameters lead to initial drape shapes that are obviously different from the GT, but are adjusted toward the GT after training.

\begin{figure*}[tb]
    \centering
    \includegraphics[width=\textwidth]{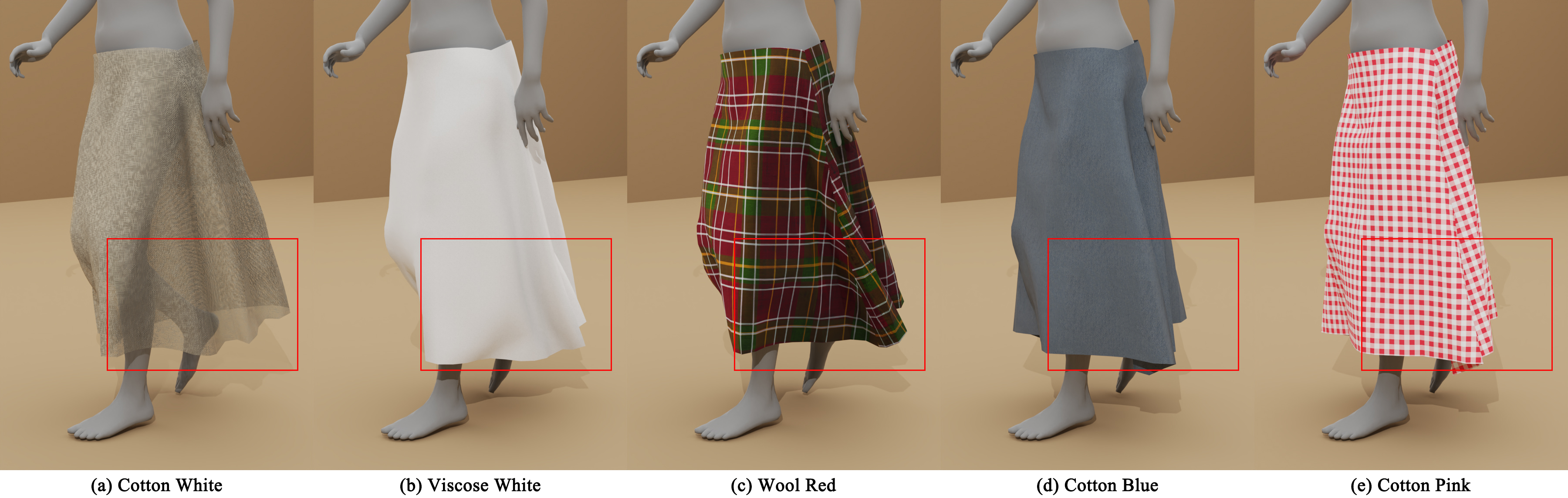}
    \caption{The digitalized cloths exhibit different mechanical characteristics in the walking motion.}
    \label{fig:comp_walk}
\end{figure*}

\begin{figure*}[tb]
    \centering
    \includegraphics[width=\textwidth]{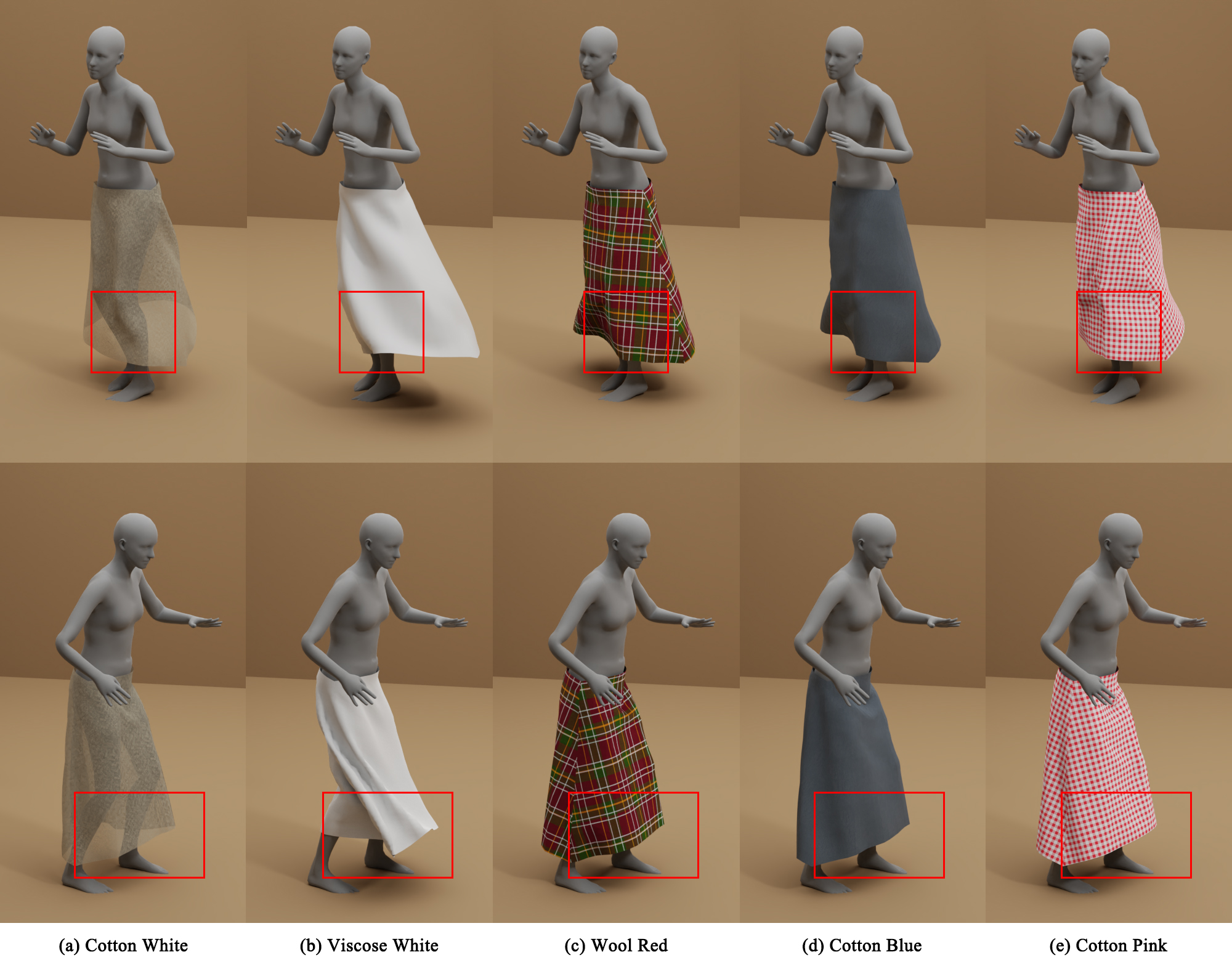}
    \caption{The digitalized cloths exhibit different mechanical characteristics in the dancing motion.}
    \label{fig:comp_dance}
\end{figure*}

\begin{figure*}[tb]
    \centering
    \includegraphics[width=\textwidth]{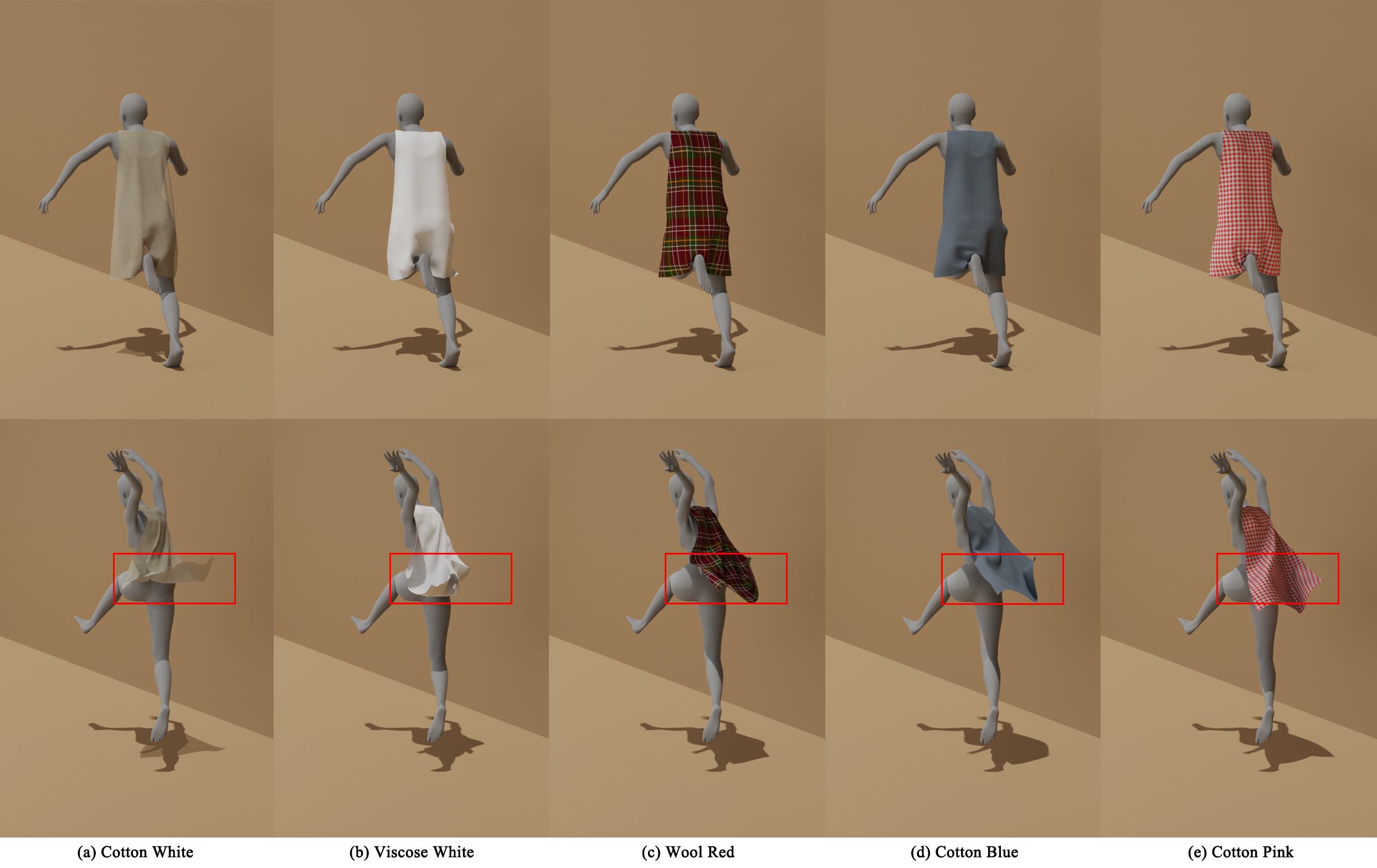}
    \caption{The digitalized cloths exhibit different mechanical characteristics in the jumping motion.}
    \label{fig:comp_jump}
\end{figure*}

\begin{figure}[tb]
    \centering
    \includegraphics[width=.47\textwidth]{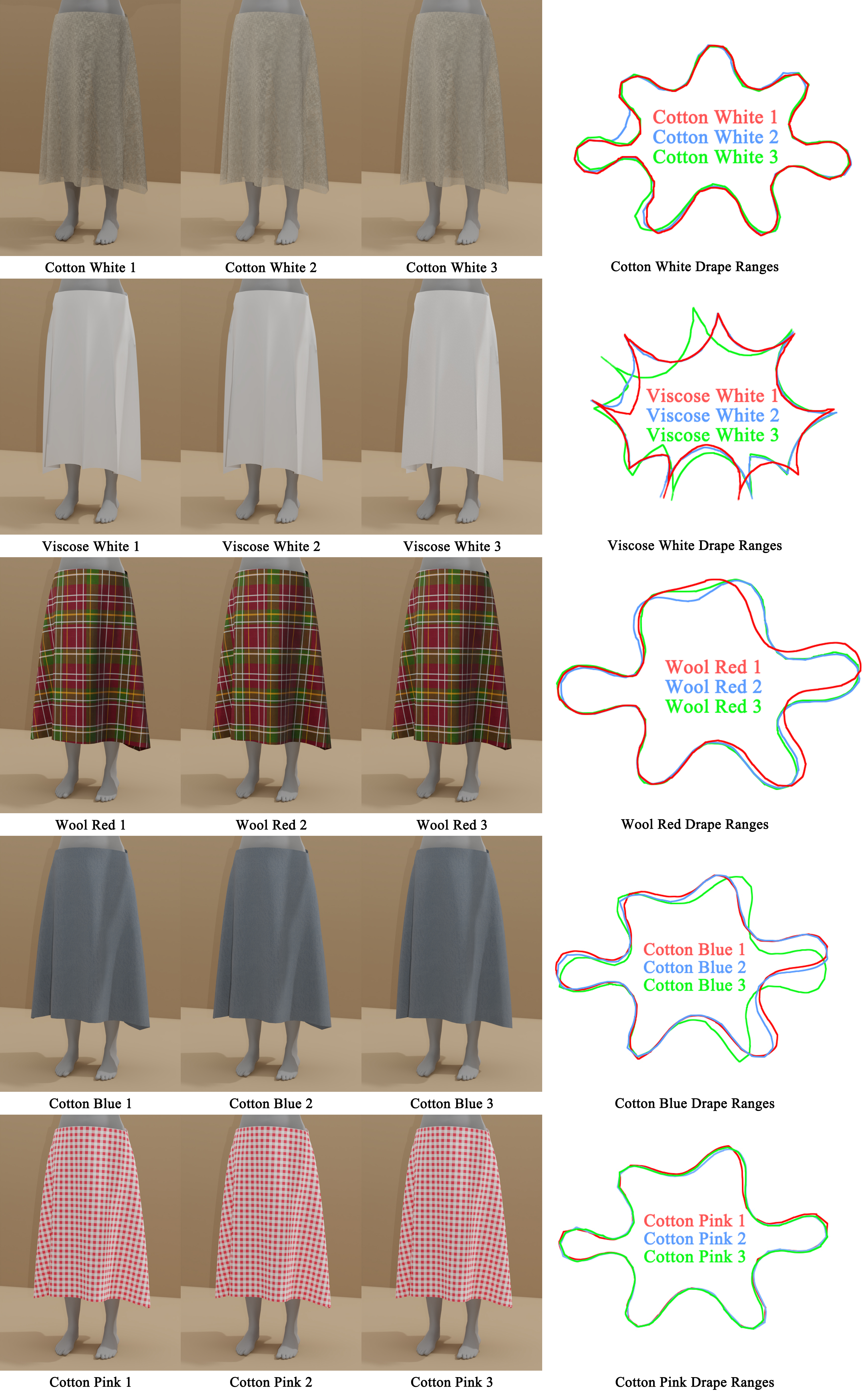}
    \caption{Given the digitialized cloths, our BDP model can simulate the skirts made from these cloths and reflect cloth material heterogeneity and draping stochasticity.}
    \label{fig:quali_comp}
\end{figure}

The digitalized cloths can be applied into garment simulation and obviously reflect different cloth mechanical characteristics in various motions (as shown in \cref{fig:comp_walk,fig:comp_dance,fig:comp_jump}). For example, in \cref{fig:comp_dance}, there are more folds on the Viscose White than on the Wool Red, because the former is much softer (small bending stiffness) than the latter. 

\begin{figure*}[tb]
    \centering
    \includegraphics[width=\textwidth]{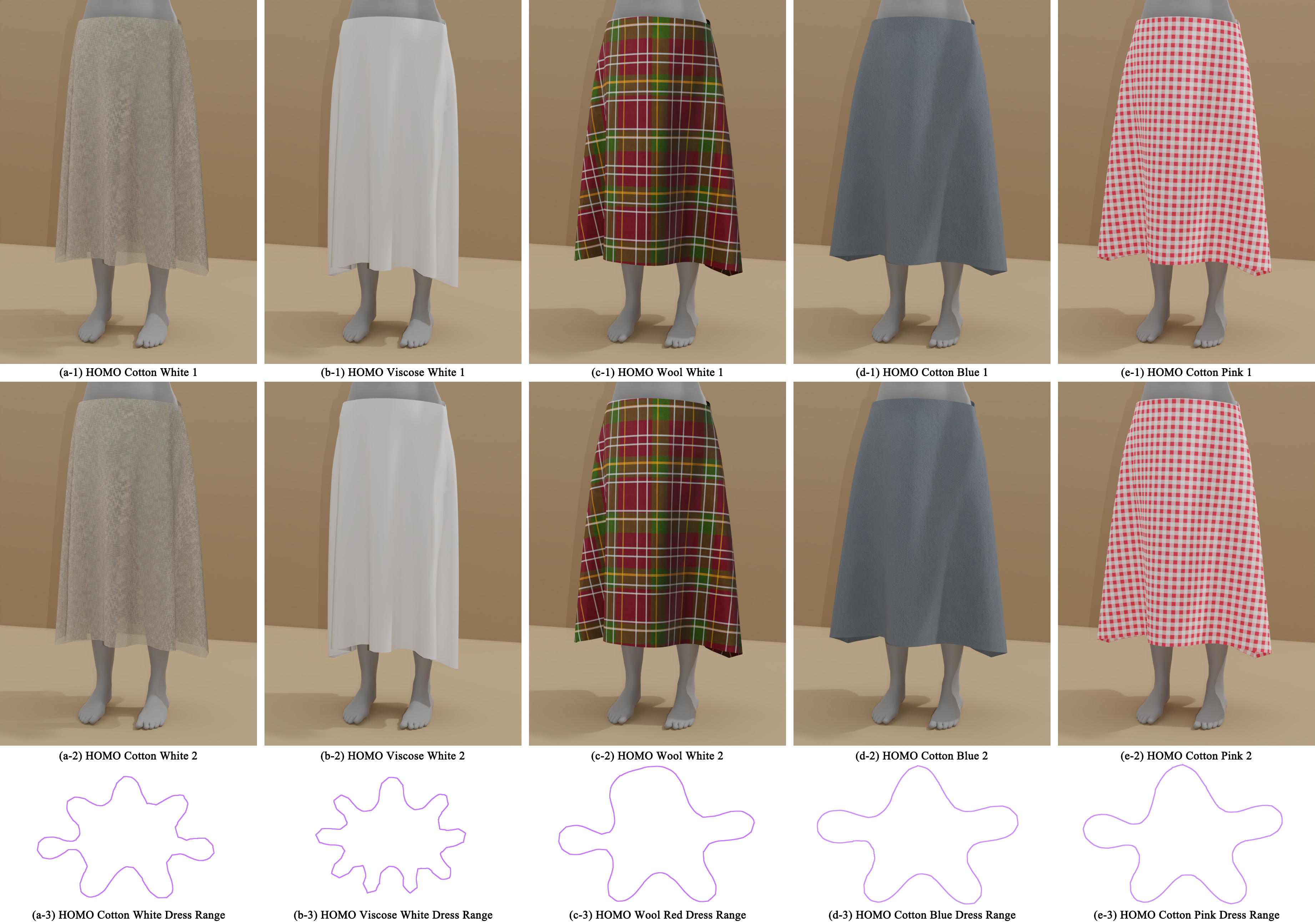}
    \caption{The clothes digitalized by the HOMO model is deterministic and has homogeneous material properties. Thus, the simulated dresses that are made from the same cloth always have the identical geometry.}
    \label{fig:dress_homo}
\end{figure*}

In addition, our BDP embeds cloth material heterogeneity and dynamics stochasticity (\cref{fig:quali_comp}). Conversely, the HOMO is unable to digitalize and simulate these cloth properties (\cref{fig:dress_homo}).

\section{More comparison}

\begin{figure}[tb]
    \centering
    \includegraphics[width=0.47\textwidth]{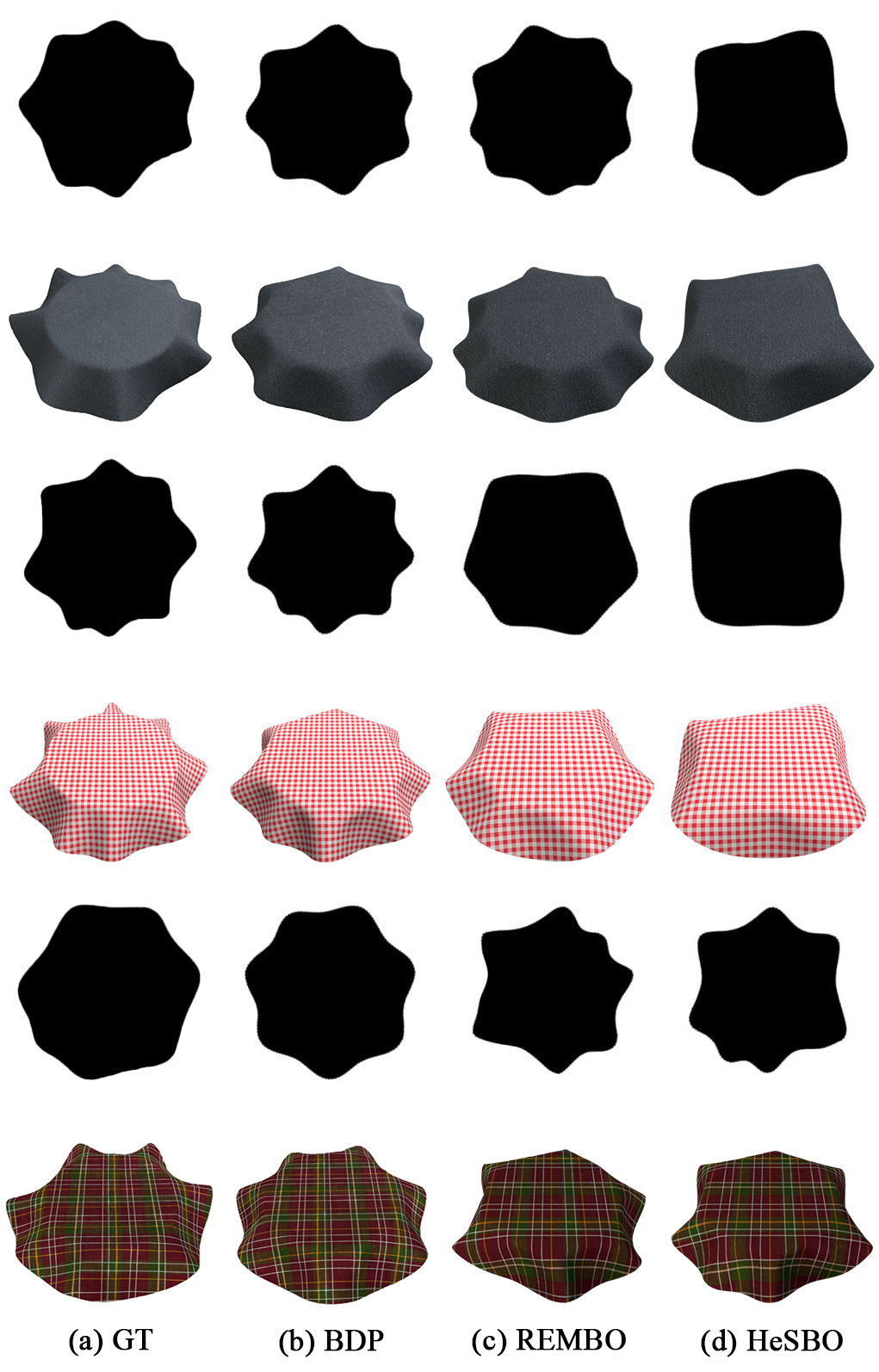}
    \caption{Comparison of ground truth draped shape and the simulated drape shapes learned by ours (b)(derivative-based), REMBO (c) and HeSBO (d)(derivative-free) optimization.}
    \label{fig:comp_opt}
\end{figure}

\cref{fig:comp_opt} provides the visual comparison between the gradient-based (our BDP) and the gradient-free optimization, \ie Bayesian Optimization (BO), method. It confirms the quantitative result given in the main paper that the BDP gains a better results with fewer optimization steps than the BO.

\section{Posterior Space}
Although we mainly focus on digitalizing cloths, the learned posterior space has other potentials. The first one is doing a quantified comparison between different cloth types. The five digitialized cloths in our experiment are representative, where we can derive some insights from the learned parameter distributions. As shown in \cref{tab:cloth_parameters}, the estimated parameters also reflect our daily qualitative observations of the cloths. For instance, the mean of the Viscose White's stretching stiffness parameters are smaller than the other cloths' because it is a knitted fabric which is soft and can be used to make T-shirt or socks which are very stretchable. By contrast, the mean of the Cotton Pink's stretching stiffness parameters are larger because it is a tightly woven fabrics which is stiff, and can be used as tablecloths. In addition, the standard deviation of the Viscose White's physical parameters are larger because its loose woven pattern tends to cause material heterogeneity and lead to dynamics stochasticity. This kind of analysis has been pursued so far in textile when it comes to Cusick drape data, but rarely with the kind of material details (reflected in \eg material heterogeneity and dynamics stochasticity) given by our BDP model.

\begin{figure*}[htb]
    \centering
    \includegraphics[width=\textwidth]{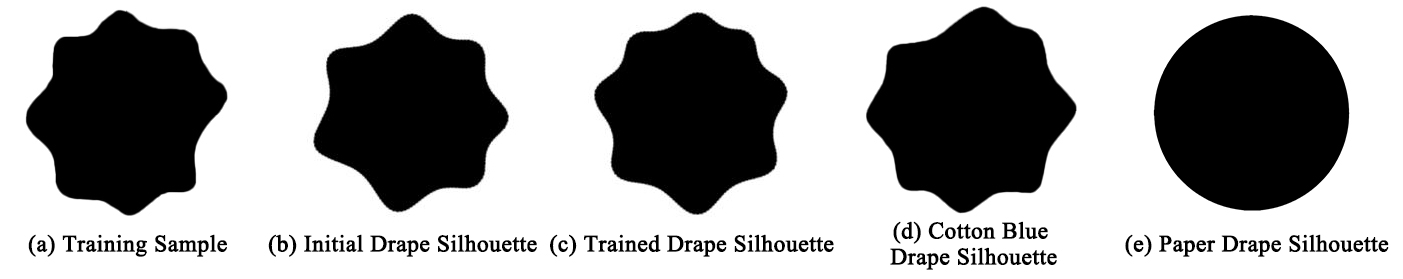}
    \caption{Through digitalization (\ie training BDP), cloth drape silhouette (b and c) is optimized toward the training sample(a). In addition, (d) shows the simulated Cusick drape silhouette of a paper-like material.}
    \label{fig:posterior_analysis}
\end{figure*}

\begin{table*}[htb]
    \centering
    \begin{tabular}{lcccccc}
    \toprule
         Cloths &  Cotton White & Cotton Blue & Viscose White & Cotton Pink & Wool Red & Sum\\
         \midrule
         Linen Green & \num{8.669e+30} & \num{1.202e+19} & \num{9.796e+30} & \num{7.307e+30} & \num{7.707e+29} & \num{2.654e+31} \\
         Paper & \num{2.253e+38} & \num{2.253e+38} & \num{2.254e+38} & \num{2.252e+38} & \num{2.253e+38} & \num{1.126e+39}\\
         \bottomrule
    \end{tabular}
    \caption{The Kullback–Leibler(KL) divergence of the estimated parameters from the Linen Green sample and the paper parameters to our selected five representative cloths' parameters probability distributions. The smaller the KL divergence is, the closer the parameters are to the distributions. Thus, (1) the ``Linen Green'' is more like Cotton Blue among the five kinds of cloths and (2) the ``Linen Green'' is more like fabric than paper.}
    \label{tab:posterior}
\end{table*}

Furthermore, enabled by BDP, we could learn all cloth types in our dataset or simply learn a subset of representative ones, then combine the learned parameter distributions into a Gaussian Mixture Model (GMM) to represent the material space of all common fabrics. This GMM could then be employed for other applications such as detecting whether a tested material is likely to be a common fabric, or even quality control in detecting anomaly material \eg due to manufacturing defects. This could be done by first learning the material, then compute the likelihood of the estimated material parameters in the GMM. Although this is way out of the scope of this paper, we can still give a simple illustrative example. We select a sample from a cloth type that is different from the five types for the learning so far, refer to it as ``Linen Green'', and digitalize it by our BDP model (\cref{fig:posterior_analysis} (a-c)). To make the comparison more visual, we make a paper-like material which shows obviously different drape shape (as shown in \cref{fig:posterior_analysis} (e)). Then we compare the likelihoods of their material parameters in the GMM consisting of the five Gaussians learned in our experiments. However, both likelihoods are very low, despite the likelihood of Linen Green is several magnitudes higher than paper-like. This suggests that only five types are not sufficient to capture the full common cloth material space. Learning more cloth types in our dataset will mitigate this issue. But we can still compute similarities between the newly learned distributions and the original five types, measured by the Kullback-Leibler (KL) Divergence . The results are shown in \cref{tab:posterior}. Interestingly, Linen Green is by far most similar to Cotton Blue compared with other types, which is consistent with qualitative observations (as shown in \cref{fig:posterior_analysis} (d)). In contrast, paper-like is dissimilar to every material, which is expected. In future, we will explore what materials need to be learned and incorporated into this GMM model so that the posterior distribution captures better the full material parameter space.


\section{Force Models}

\begin{figure*}[bt]
    \centering
    \includegraphics[width=0.9\textwidth]{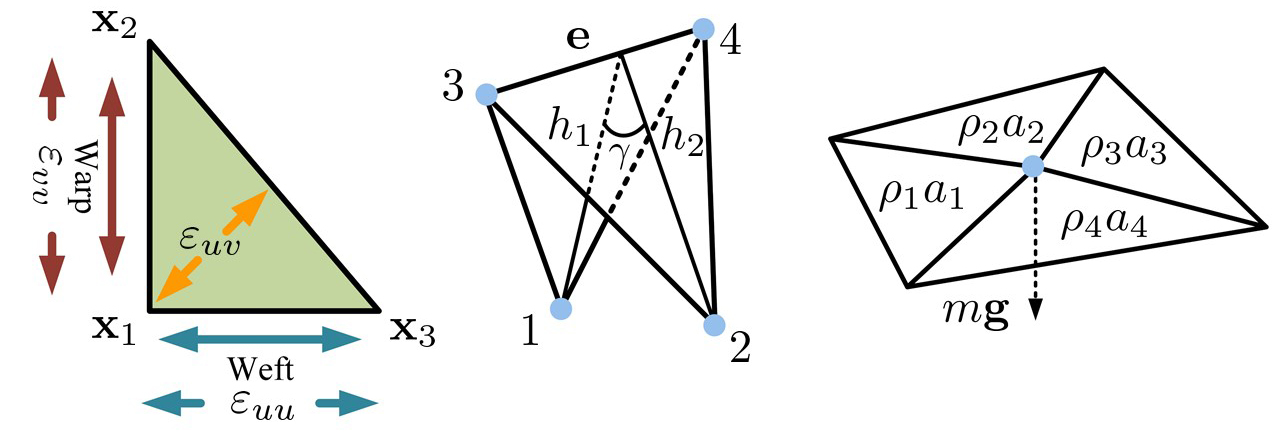}
    \caption{Left: cloth weft and warp directions (or course and wale directions in knitted cloths) and three strain directions in a triangle. Middle: bending force between two adjacent triangles. Right: triangle mass and gravity.}
    \label{fig:forces}
\end{figure*}

In this section, we give more details about our cloth model. A cloth sample is modeled as a circular triangular mesh consisting of $v$ vertices. Its state $\mathcal{S}$ is defined by its vertices position, $\mathbf{x} \in \mathbb{R}^{3 \times v}$, and velocity, $\dot{\mathbf{x}} \in \mathbb{R}^{3 \times v}$, in Cartesian coordinates. Therefore, a cloth motion is represented by a state sequence: $\mathcal{S}_{0:n} = \{\mathcal{S}_t : t \in \mathbb{Z}^{+}; t \leq n\}$, which denotes the position and velocity change over time. A physics-based cloth simulator aims to simulate a cloth motion by recurrently predicting its future state $\mathcal{S}_{t+1} = \{\mathbf{x}_{t+1}, \dot{\mathbf{x}}_{t+1} \}$ given the current state $\mathcal{S}_{t} = \{\mathbf{x}_{t}, \dot{\mathbf{x}}_{t} \}$:
\begin{gather}
    \mathbf{x}_{t+1} = \mathbf{x}_{t} + h \dot{\mathbf{x}}_{t} \\
    \dot{\mathbf{x}}_{t+1} = \dot{\mathbf{x}}_{t} + h \ddot{\mathbf{x}}_{t}
\end{gather}
where $h$ is the time step size (time lapse between every two consecutive states) and the second-order time derivative, $\ddot{\mathbf{x}}_{t}$, is vertices acceleration. To gain high simulation stability, implicit Euler method \cite{baraff1998large} is commonly used:
\begin{gather}
    \label{eq:imp_pos} \mathbf{x}_{t+1} = \mathbf{x}_{t} + h \dot{\mathbf{x}}_{t+1} \\
    \label{eq:imp_vel} \dot{\mathbf{x}}_{t+1} = \dot{\mathbf{x}}_{t} + h \ddot{\mathbf{x}}_{t+1} 
\end{gather}
According to Newton's Second law, we have
\begin{equation}
    \mathbf{F} = \mathbf{Ma} = \mathbf{M\ddot{x}}
\end{equation}
where $\mathbf{M}$ is the general mass matrix and $\mathbf{F}$ is the resultant force which is the combination of internal and external forces. In our differentiable cloth simulator, these forces are decided by cloth sample's current state, so we can define:
\begin{equation}
    \mathbf{F}_t = \mathbf{f}(\mathcal{S}_t) = \mathbf{f}(\mathbf{x}_t, \dot{\mathbf{x}}_t)
\end{equation}
where $\mathbf{f}$ denotes a general function, which must be differentiable, takes as input the current state, $\mathcal{S}_t$, and outputs the resultant force. Through Taylor approximation\cite{baraff1998large}, \cref{eq:imp_pos} and \cref{eq:imp_vel} are converted to the governing equation of the physical system: 
\begin{equation}
    \left( \mathbf{M} - h \frac{\partial \mathbf{f}}{\partial \dot{\mathbf{x}}} - h^2 \frac{\partial \mathbf{f}}{\partial \mathbf{x}} \right) \Delta \dot{\mathbf{x}} = h \left( \mathbf{F}_{t} + h \frac{\partial \mathbf{f}}{\partial \mathbf{x}} \dot{\mathbf{x}}_{t}\right)
\label{eq:motion}
\end{equation}
which needs to be solved to calculate $\Delta \dot{\mathbf{x}}$ and update the cloth sample's state:
\begin{gather}
    \dot{\mathbf{x}}_{t+1} = \dot{\mathbf{x}}_{t} + \Delta \dot{\mathbf{x}} \\
    \mathbf{x}_{t+1} = \mathbf{x}_{t} + h \dot{\mathbf{x}}_{t+1} 
\end{gather}
In our differentiable cloth simulator, the resultant force is defined as: $\mathbf{F} = \mathbf{F}_{stretch} + \mathbf{F}_{bend} + \mathbf{F}_{gravity} + \mathbf{F}_{handle}$. The stretching force~\cite{volino2009simple} on a face $j$ is:
\begin{equation}
    \mathbf{F}_{stretch}^{(j)} = - A^{(j)}  
    \left(
        \sum_{m \in (uu, vv, uv)} \sigma_m^{(j)}
        \left(
            \frac{\partial \varepsilon_m^{(j)}}{\partial \mathbf{x}_i}
        \right)
    \right)
\end{equation}
where $\varepsilon_m^{(j)}$ denotes stretching strain and the $\mathbf{x}_i$ are the three vertices of the face. The stretching stresses $\sigma_m^{(j)}$ is
\begin{equation}
    \begin{bmatrix}
        \sigma_{uu}^{(j)} \\
        \sigma_{vv}^{(j)} \\
        \sigma_{uv}^{(j)}
    \end{bmatrix} = 
    \begin{bmatrix}
        c_{11}^{(j)} & c_{12}^{(j)} & 0 \\
        c_{12}^{(j)} & c_{22}^{(j)} & 0 \\
        0            & 0            & c_{33}^{(j)}
    \end{bmatrix}
    \begin{bmatrix}
        \varepsilon_{uu}^{(j)} \\
        \varepsilon_{vv}^{(j)} \\
        \varepsilon_{uv}^{(j)}
    \end{bmatrix} = \mathbf{C}^{(j)} \boldsymbol{\varepsilon}^{(j)}
    \label{eq:stretching}
\end{equation}
where $c_{11}^{(j)}$, $ c_{12}^{(j)}$, $c_{22}^{(j)}$, and $c_{33}^{(j)}$ are the stretching stiffness in weft/course direction, the stretching stiffness in warp/wale direction, Poisson's ratio, and shearing stiffness. The subscripts $uu$, $vv$, and $uv$ denote cloths weft/course, warp/wale, and diagonal directions respectively (\cref{fig:forces} Left). $\boldsymbol{\varepsilon} = [\varepsilon_{uu}^{(j)}, \varepsilon_{vv}^{(j)}, \varepsilon_{uv}^{(j)}]$ is the Voige form strain tensor. The bending force around a bending edge $w$ is defined as 
\begin{equation}
    \mathbf{F}_{bend}^{(w)} = \mathbf{B}^{(w)} \frac{|\mathbf{e}^{(w)}|}{\psi_1^{(w)} + \psi_2^{(w)}} \sin(\frac{\gamma^{(w)}}{2} - \frac{\bar{\gamma}^{(w)}}{2} )  u_i
    \label{eq:bending}
\end{equation}
where $\mathbf{B}^{(w)}$ is the bending edge $w$'s bending stiffness, $|\mathbf{e}^{(w)}|$ is bending edge's rest length, $\psi_1^{(w)}$ and $\psi_2^{(w)}$ are the heights of the two adjacent triangular face, $\gamma^{(w)}$ and $\bar{\gamma^{(w)}}$ is the current and predefined rest dihedral angles between the edge's two adjacent faces (\cref{fig:forces} Middle). 

Material non-linearity means the material stiffness changes with deformation magnitude non-linearly. Anisotropy refers to the varied material stiffness in different deformation directions. To encode cloth material non-linearity and anisotropy, our model adopts the piecewise linear physical models in \cite{wang2011data} where the stretching stiffness and bending stiffness are defined as two high-dimensional matrices: $\mathbf{C} \in \mathbb{R}^{6 \times 4}$ and $\mathbf{B} \in \mathbb{R}^{3 \times 5}$. Then, local stretching stiffness and bending stiffness are sampled from $\mathbf{C}$ and $\mathbf{B}$ according to the mesh's local deformation and geometry. Wang et al.\cite{wang2011data} model cloths as continuum elastic shells so a stretching deformation can be described by the Green-Lagrangian strain tensor \cite{bonet2008nonlinear}, which can be re-parameterized by Eigen Decomposition:
\begin{align}
    &2
    \begin{bmatrix}
        \varepsilon_{uu}^{(j)} & \varepsilon_{uv}^{(j)} \\
        \varepsilon_{uv}^{(j)} & \varepsilon_{vv}^{(j)} \\
    \end{bmatrix} = \notag \\
    & (\mathbf{R}^{(j)}_\varphi)^{\top}
    \begin{bmatrix}
        (\lambda_{max}^{(j)} + 1)^2 - 1 & 0 \\
        0 & (\lambda_{min}^{(j)} + 1)^2 - 1 \\
    \end{bmatrix}
    \mathbf{R}_\varphi^{(j)}
\end{align}
where the eigenvalues indicate the stretching deformation magnitude and $\mathbf{R}_\varphi^{(j)}$ is a rotation matrix that indicates the direction of the stretching deformation. $\lambda_{min}^{(j)}$ can be ignored because they find it has less influence on the stretching stiffness. The rotation matrix is decided by the bias angle, $\varphi$, between a cloth sample's rest warp-weft coordinate system and its deformed local coordinate system \cite{peng2005continuum}. This way, the stretching non-linearity and anisotropy can be encoded as the stiffness which changes with parameters in the 2D space spanned by $\lambda_{max}^{(j)}$ and $\varphi$. As in \cite{wang2011data}, we sample 6 data points (the 6 rows in the matrix $\mathbf{C}$) in the polar space spanned by $\lambda_{max}$ and $\varphi$ where each data point contains $c_{11}$, $c_{12}$, $c_{22}$, and $c_{33}$ which compose the 4 columns in the matrix $\mathbf{C}$. (We ignore the face index superscript, $(j)$, to denote the general form.)

\begin{figure}[tb]
    \centering
    \begin{tikzpicture}[scale=2]
        \draw[blue, very thick] (0,0) coordinate(B) -- (0.3,1.0) coordinate(C);
        \draw[blue, very thick] (1.1,-0.1) coordinate(E) -- (0.3,1.0);
        \draw (1.1,-0.1) -- (0,0);
        \draw (0,0) -- (-0.9,-0.1);
        \draw (0.3,1.0) -- (-0.9,-0.1);
        \draw (0.3,1.0) -- (1.6,0.9);
        \draw (1.6,0.9) -- (1.1,-0.1);
        \draw[->, gray, very thin] (-0.3,0) -- (0.5,0) node[black,above] {weft} coordinate(A);
        \draw[->, gray, very thin] (0,-0.3) -- (0,0.5) node[black,left] {warp};
        \draw[->, gray, very thin] (0.6,-0.1) -- (1.6,-0.1) node[black,above] {weft} ;
        \draw[->, gray, very thin] (1.1,-0.3) -- (1.1,0.4) node[black,right] {warp} coordinate(D);
        \draw pic [draw=green!50!black, fill=green!20, angle radius=5mm, "${\scriptstyle \beta_1}$"] {angle = A--B--C};
        \draw pic [draw=green!50!black, fill=green!20, angle radius=7mm, "${\scriptstyle \beta_2}$"] {angle = D--E--C};
    \end{tikzpicture}
    \caption{The bending bias angles, $\beta_1$ and $\beta_2$, of the two bending edges (blue).}
    \label{fig:bending_bias_angle}
\end{figure}
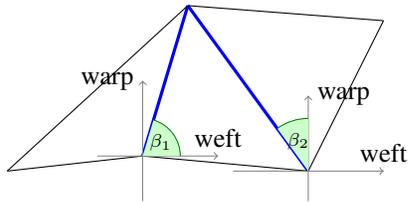

To model non-linear bending stiffness, the variables in \cref{eq:bending} can represent the bending deformation so we define a parameter $\alpha$:
\begin{equation}
    \alpha^{(w)} = \frac{\sin(\frac{\gamma^{(w)}}{2} - \frac{\bar{\gamma}^{(w)}}{2})}{h_1^{(w)} + h_2^{(w)}}
\end{equation}
which is related to the curvature. To model the bending anisotropy, we define another parameter called bending bias angle, i.e. the angle between a bending edge and cloth warp-weft coordinate system's axes, which indicates bending deformation direction (shown in \cref{fig:bending_bias_angle}). Therefore, the bending non-linearity and anisotropy can be encoded as the stiffness which changes with the parameter in a polar space spanned by $\alpha$ and bending bias angle. We sample 5 $\alpha$'s and 3 bending bias angles ($0^{\circ}$, $45^{\circ}$, and $90^{\circ}$) which are the 5 columns and the 3 rows of bending stiffness matrix $\mathbf{B}$. 

Finally, to model the material heterogeneity, each mesh face and bending edge are associated with a $\mathbf{C}$ and $\mathbf{B}$. Therefore, for a cloth consisting of $f$ faces and $e$ bending edges, the learnable parameters are $f$ stretching matrices and $e$ bending stiffness matrices (the HETER model in our experiments). To our Bayesian differentiable cloth simulator, each $\mathbf{C}$ and $\mathbf{B}$ are sampled from the variational distribution $q_{\theta}(\tau)$. Its learnable parameter is the distribution parameter $\theta$ of $q_{\theta}(\tau)$. As we assume $q_{\theta}(\tau)$ is distributed as a Gaussian, $\theta$ consists of the means and the variances of the stretching stiffness and bending stiffness: $2 \times 4 \times 6 + 2 \times 3 \times 5 = 78$ learnable parameters. 

The gravity, $\mathbf{F}_{gravity}$, is calculated on every face and evenly divided by its three vertices. Therefore, the gravity on the \textit{k}th vertex is 
\begin{equation}
    \mathbf{F}_{gravity}^{(k)} 
    = m^{(k)} \textbf{g} 
    = \sum_{j=0}^{n_j^{(k)}} \frac{1}{3} \rho^{(j)} A^{(j)} \textbf{g}
\end{equation}
where $\rho^{(j)}$ is the \textit{j}th face's area density and $\textbf{g} = [0.0, 0.0, -9.8]^{\top} m/s^2$. $n_j^{(k)}$ is number of the adjacent faces of vertex \textit{k} (\cref{fig:forces} Right). The handle force, $\mathbf{F}_{handle}$, is used to pin and support a cloth sample, i.e. simulating the inner support panel. To each vertex whose distance to the center of the cloth is smaller than 9cm, the handle force is computed as:
\begin{equation}
    \mathbf{F}_{handles}^{(k)} = k_h \mathbf{I}_3 (\mathbf{x}^{(k)} - \mathbf{\bar{x}}^{(k)})
\end{equation}
where $\bar{\mathbf{x}}^{(k)}$ denotes the $k$th vertex's anchor position where the vertex should be fixed and $k_h$ is the handle stiffness.

\section{Derivatives of the Simulator}

Now we have a fully differentiable cloth simulator. We then compute the loss $\mathcal{L}$ that indicates the difference between the predicted and ground truth cloth states. The loss gradients with respect to the parameters $\frac{\partial \mathcal{L}}{\partial w}$ can help learn the right physics parameters via back-propagation. For simplicity, we use $\mathbf{A y = b}$ to represent Equation \ref{eq:motion}. The differential of $\mathbf{A y = b}$ is~\citep{magnus_matrix_2019}:
\begin{equation}
    \label{eqn:app_der_1}
     \mathbf{A} \mathsf{d} \mathbf{y} = \mathsf{d} \mathbf{b} - \mathsf{d}\mathbf{A} \mathbf{y}
\end{equation}
We can form the Jacobians of $\mathbf{y}$ with respect to $\mathbf{A}$ or $\mathbf{b}$ with Equation \ref{eqn:app_der_1}. For example, to compute the $\frac{\partial \mathbf{y}}{\partial \mathbf{A}}$, we need to set $\mathsf{d}\mathbf{A} = \mathbf{I}$ and $\mathsf{d}\mathbf{b}=\mathbf{0}$, then solve the equation and the result is $\frac{\partial \mathbf{y}}{\partial \mathbf{A}}$. As pointed out by \citet{amos2017optnet}, it is unnecessary to explicitly compute these Jacobians in back-propagation. We want to compute the product of the vector passed from back-propagation, $\frac{\partial \mathcal{L}}{\partial \mathbf{y}}$ and the Jacobians of $\mathbf{y}$, i.e.$\frac{\partial \mathcal{L}}{\partial \mathbf{y}} \frac{\partial \mathbf{y}}{\partial \mathbf{A}}$ and $\frac{\partial \mathcal{L}}{\partial \mathbf{y}} \frac{\partial \mathbf{y}}{\partial \mathbf{b}}$. Assume $\mathbf{A} \in \mathbb{R}^{3\times3}$, $\mathbf{y} \in \mathbb{R}^3$, and $\mathbf{b} \in \mathbb{R}^3$, then 
\begin{align}
    \label{eqn:der_2}
    &\frac{\partial \mathcal{L}}{\partial \mathbf{b}}
    = \frac{\partial \mathcal{L}}{\partial \mathbf{y}} \frac{\partial \mathbf{y}}{\partial \mathbf{b}} \notag \\
    &= \left(
    \begin{pmatrix}
        \frac{\partial \mathcal{L}}{\partial \mathbf{y}_1} &
        \frac{\partial \mathcal{L}}{\partial \mathbf{y}_2} &
        \frac{\partial \mathcal{L}}{\partial \mathbf{y}_3}
    \end{pmatrix}
    \begin{pmatrix}
        \frac{\partial \mathbf{y}_1}{\partial \mathbf{b}_1} & \frac{\partial \mathbf{y}_1}{\partial \mathbf{b}_2} & \frac{\partial \mathbf{y}_1}{\partial \mathbf{b}_3}\\
        \frac{\partial \mathbf{y}_2}{\partial \mathbf{b}_1} & \frac{\partial \mathbf{y}_2}{\partial \mathbf{b}_2} & \frac{\partial \mathbf{y}_2}{\partial \mathbf{b}_3}\\
        \frac{\partial \mathbf{y}_3}{\partial \mathbf{b}_1} & \frac{\partial \mathbf{y}_3}{\partial \mathbf{b}_2} & \frac{\partial \mathbf{y}_3}{\partial \mathbf{b}_3}
    \end{pmatrix}
    \right)^{\top} 
\end{align}
As
\begin{equation}
    \label{eqn:der_3}
    \frac{\partial \mathbf{y}_1}{\partial \mathbf{b}_1} 
    = \frac{ \partial
    \; (\mathbf{A}^{-1})_{1,1}\mathbf{b}_1
    + (\mathbf{A}^{-1})_{1,1}\mathbf{b}_2
    + (\mathbf{A}^{-1})_{1,1}\mathbf{b}_3}
    { \partial \mathbf{b}_{1}}
    = \mathbf{A}^{-1}_{1,1} \nonumber
\end{equation}
and similarly for $\frac{\partial \mathbf{y}_i}{\partial \mathbf{b}_j}$, Equation \ref{eqn:der_2} can be represented as: 
\begin{align}
\label{eqn:der_4}
    &\left(
    \begin{pmatrix}
        \frac{\partial \mathcal{L}}{\partial \mathbf{y}_1} &
        \frac{\partial \mathcal{L}}{\partial \mathbf{y}_2} &
        \frac{\partial \mathcal{L}}{\partial \mathbf{y}_3}
    \end{pmatrix}
    \begin{pmatrix}
        (\mathbf{A}^{-1})_{1,1} & (\mathbf{A}^{-1})_{1,2} & (\mathbf{A}^{-1})_{1,3} \\
        (\mathbf{A}^{-1})_{2,1} & (\mathbf{A}^{-1})_{2,2} & (\mathbf{A}^{-1})_{2,3} \\
        (\mathbf{A}^{-1})_{3,1} & (\mathbf{A}^{-1})_{3,2} & (\mathbf{A}^{-1})_{3,3}
    \end{pmatrix}
    \right)^{\top} \notag \\
    &= (\mathbf{A}^{-1})^{\top} \frac{\partial \mathcal{L}}{\partial \mathbf{y}}
\end{align}
After computing $\frac{\partial \mathcal{L}}{\partial \mathbf{b}}$, we need to compute $\frac{\partial \mathcal{L}}{\partial \mathbf{A}}$. The $\mathsf{}\mathbf{b}$ in Equation \ref{eqn:app_der_1} can be set to 0 because it is irrelevant when computing $\frac{\partial \mathcal{L}}{\partial \mathbf{A}}$. Then we have
\begin{equation}
    \mathbf{A} \mathsf{d} \mathbf{y} = - \mathsf{d}\mathbf{A} \mathbf{y}
\end{equation}
The derivative of $\mathbf{y}$ with respect to $\mathbf{A}_{i,j}$, the entry in the $i$th row and $j$th column of the matrix $\mathbf{A}$, is
\begin{equation}
    \frac{\partial \mathbf{y}} {\partial \mathbf{A}_{i,j}} = \mathbf{A}^{-1}
    \begin{pmatrix}
        \mathbf{0} \\ -\mathbf{y}_j \\ \mathbf{0}
    \end{pmatrix}
\end{equation}
According to chain rule, 
\begin{align}
    \frac{\partial \mathcal{L}}{\partial \mathbf{A}_{i,j}} 
    &= \frac{\partial \mathcal{L}}{\partial \mathbf{y}} 
    \frac{\partial \mathbf{y}}{\partial \mathbf{A}_{i,j}} 
    = \frac{\partial \mathcal{L}}{\partial \mathbf{b}}^{\top}\mathbf{A}\mathbf{A}^{-1}
    \begin{pmatrix}
        \mathbf{0} \\ -\mathbf{y}_j \\ \mathbf{0}
    \end{pmatrix} \notag \\
    &= - \left(\frac{\partial \mathcal{L}}{\partial \mathbf{b}}\right)_i
    \mathbf{y}_j
\end{align}
The more general form is 
\begin{equation}
    \frac{\partial \mathcal{L}}{\partial \mathbf{A}} = -
    \frac{\partial \mathcal{L}}{\partial \mathbf{b}} \mathbf{y}^{\top}
\end{equation}

{
    \small
    \bibliographystyle{ieeenat_fullname}
    \bibliography{supplement}
}